\documentclass[letterpaper]{article} 
\usepackage{aaai2026}  
\usepackage{times}  
\usepackage{helvet}  
\usepackage{courier}  
\usepackage[hyphens]{url}  
\usepackage{graphicx} 
\usepackage{natbib}  
\usepackage{caption} 
\usepackage{algorithm}
\usepackage{algorithmic}
\usepackage{framed}
\usepackage{booktabs} 
\usepackage{multirow}
\usepackage{makecell}
\usepackage{verbatim} 
\usepackage{comment}

\usepackage{newfloat}
\usepackage{listings}
\usepackage[most]{tcolorbox}   
\tcbset{enhanced,
        arc=2pt,                       
        boxrule=0.7pt,                 
        colframe=black,                
        colback=white,                 
        fonttitle=\bfseries,           
        colbacktitle=black!10,         
        coltitle=black,                
        left=3pt,right=3pt,top=3pt,bottom=3pt   
}

\definecolor{shadecolor}{gray}{0.9}
\DeclareCaptionStyle{ruled}{labelfont=normalfont,labelsep=colon,strut=off} 
\floatstyle{ruled}
\newfloat{listing}{tb}{lst}{}
\floatname{listing}{Listing}
\title{PerPilot: Personalizing VLM-based Mobile Agents via Memory and Exploration}
\author{
    Xin Wang\textsuperscript{\rm 1 \dag}, Zhiyao Cui\textsuperscript{\rm 1, 2 \dag},Hao Li\textsuperscript{\rm 1, 2 \dag}, 
    Ya Zeng\textsuperscript{\rm 1}, Chenxu Wang\textsuperscript{\rm 2}, 
    Ruiqi Song\textsuperscript{\rm 1}, Yihang Chen\textsuperscript{\rm 3}, 
    Kun Shao\textsuperscript{\rm 3}, Qiaosheng Zhang\textsuperscript{\rm 2}, 
    Jinzhuo Liu\textsuperscript{\rm 4}, Siyue Ren\textsuperscript{\rm 1, 2}, 
    Shuyue Hu\textsuperscript{\rm 2} and Zhen Wang\textsuperscript{\rm 1}
}
\affiliations{
    \textsuperscript{\rm 1}Northwestern Polytechnical University\\
    \textsuperscript{\rm 2}Shanghai Artificial Intelligence Laboratory\\
    \textsuperscript{\rm 3}Huawei Noah's Ark Lab\\
    \textsuperscript{\rm 4}Yunnan University\\

    xinwang\_nwpu@outlook.com, \{zhiyao, li.hao, rensiyue\}@mail.nwpu.edu.cn
}

\usepackage{bibentry}
\nocopyright
\begin{document}

\maketitle

\begin{abstract}
\renewcommand{\thefootnote}{\fnsymbol{footnote}}
\footnotetext[2]{Co-first, equal contributions}
Vision language model (VLM)-based mobile agents show great potential for assisting users in performing instruction-driven tasks. However, these agents typically struggle with personalized instructions---those containing ambiguous, user-specific context---a challenge that has been largely overlooked in previous research. In this paper, we define personalized instructions and introduce \emph{PerInstruct}, a novel human-annotated dataset covering diverse personalized instructions across various mobile scenarios. Furthermore, given the limited personalization capabilities of existing mobile agents, we propose \emph{PerPilot}, a plug-and-play framework powered by large language models (LLMs) that enables mobile agents to autonomously perceive, understand, and execute personalized user instructions. PerPilot identifies personalized elements and autonomously completes instructions via two complementary approaches: memory-based retrieval and reasoning-based exploration. Experimental results demonstrate that PerPilot effectively handles personalized tasks with minimal user intervention and progressively improves its performance with continued use, underscoring the importance of personalization-aware reasoning for next-generation mobile agents. The dataset and code are available at: 
https://github.com/xinwang-nwpu/PerPilot


\end{abstract}

\begin{figure*}[ht]
  \centering
  \includegraphics[width=0.97\linewidth]{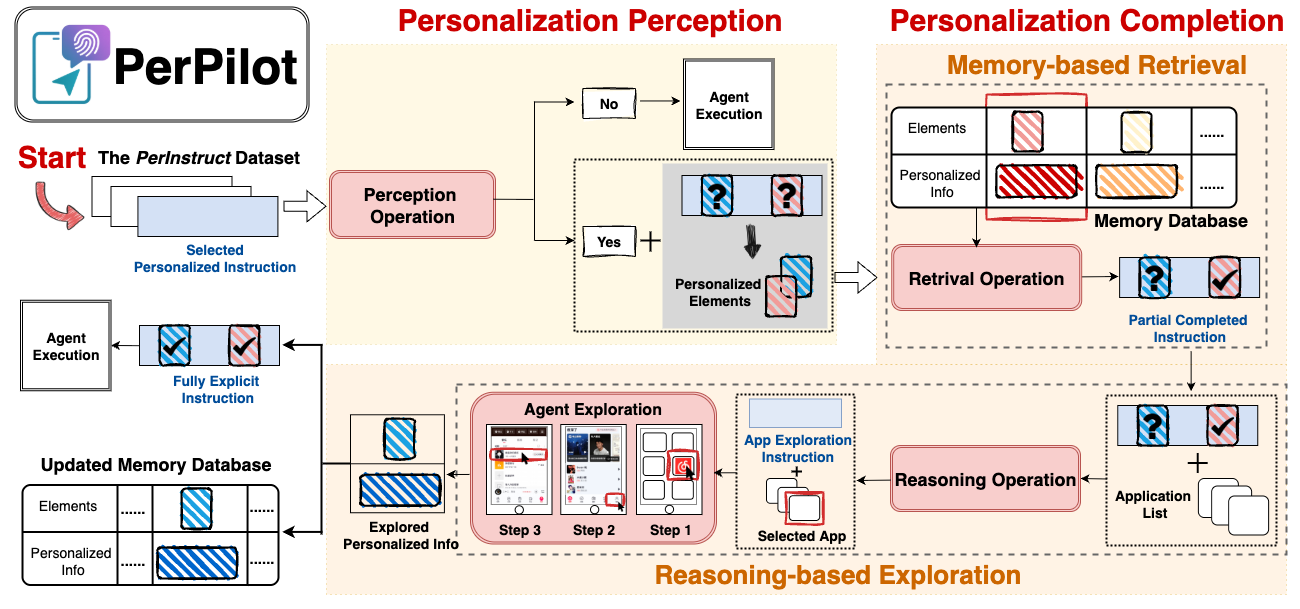}
  \caption{\emph{PerPilot}: our plug-and-play LLM-powered framework enabling mobile agents to perceive, understand, and execute personalized user instructions. First, by the \emph{Personalization Perception} module, agents identify personalized instructions and extract user-specific elements. Next, agents retrieve personalized information via memory-based retrieval or reasoning-based exploration using the \emph{Personalization Completion} module, generating actionable instructions. Finally, by integrating with existing VLM-based mobile agent systems, agents autonomously complete tasks by executing these clarified instructions.}
  \label{fig: framework}
\end{figure*}

\section{Introduction}

Recent advancements in foundation models have accelerated the rapid progress in mobile agents based on vision language models (VLMs) or multimodal large language models (MLLMs), enabling seamless perception and interaction across web, desktop, and mobile environments~\cite{wang2025mobile,yao2022webshop,gur2024real}. These agents have demonstrated capabilities such as recognizing objects and scenes from mobile images and executing user text or spoken instructions. They can autonomously navigate GUI interfaces by performing continuous sequences of actions from an initial screen state to task completion, effectively managing complex multi-step operations with minimal human oversight~\cite{deng2024mobile,qin2025ui,wang2025mobile}.


However, despite these advancements, current mobile agents exhibit a critical limitation: they lack the capability to perceive and interpret personalized user instructions---a challenge largely overlooked in previous research. Existing benchmarks predominantly focus on clearly defined, explicit tasks, such as ``Set an alarm for 7:30 am,'' neglecting the widespread occurrence of personalized instructions in everyday interactions, which inherently involve ambiguity, such as ``Play my favorite song'' or ``Set an alarm for my usual wake-up time''~\cite{fan2006personalization,riecken2000introduction,deng2024mobile}. These personalized instructions embed user-specific preferences and routines, requiring agents to first accurately perceive and understand these personalized elements before successfully executing the corresponding tasks.

To bridge this gap, we define \emph{personalized instructions} as those containing ambiguous, user-specific elements (referred to as \emph{personalized elements}) that reflect individual preferences and routines, typically stable and consistent over an extended period~\cite{kocaballi2019personalization,akasaki2024detecting}. Guided by this, we introduce \emph{PerInstruct}, the \emph{first} benchmark explicitly designed to evaluate the personalization capabilities of VLM-based mobile agents. \emph{PerInstruct} comprises $75$ personalized instructions spanning over $27$ widely-used mobile applications (apps), capturing diverse everyday scenarios. The dataset is constructed by combining authentic user-generated requests with instructions augmented through LLM-based generation. To systematically assess varying degrees of personalization complexity, we categorize the instructions into three distinct levels---\textit{simple}, \textit{normal}, and \textit{hard}---based on the quantity of personalized elements and the number of involved applications.

How can we empower VLM-based mobile agents to autonomously understand personalized user instructions? A straightforward solution is directly prompting users to clarify ambiguous instructions; however, frequent user prompts negatively impact convenience and user experience. In practice, we observe two key insights that can help address this challenge: (i) personalized information, reflecting individual preferences, tends to remain stable and consistent over extended periods; and (ii) such personalized elements are often accessible through widely-used mobile apps. For instance, the personalized instruction ``Navigate to my home'' includes the personalized element (``my home''), which typically exists in apps such as Amazon or Taobao. 

To this end, we propose \emph{PerPilot}, a plug-and-play framework powered by large language models (LLMs) that enables mobile agents to autonomously perceive and execute personalized instructions with minimal human intervention. PerPilot operates as follows: upon receiving an instruction, the \emph{Perception} module first identifies whether the instruction contains user-specific personalized elements. Subsequently, the \emph{Completion} module retrieves corresponding personalized information by initially conducting \emph{memory-based retrieval}, searching its memory for previously stored element-information pairs (e.g., mapping ``mom'' to ``Susan''). If the required information is unavailable in memory, PerPilot then performs \emph{reasoning-based exploration}, autonomously reasoning about which mobile apps likely contain the missing information and generating instructions to explore and retrieve it. Once retrieved, the information is stored in memory, and PerPilot generates a fully clarified instruction by replacing personalized elements with explicit details. Integrated with existing agent systems~\cite{wang2025mobile,zhang2025appagent,qin2025ui}, PerPilot empowers mobile agents to autonomously handle personalized user instructions. Figure~\ref{fig: framework} illustrates the PerPilot architecture.

In our experiments, we evaluate PerPilot’s personalization capabilities by integrating it into three leading mobile-agent systems: AppAgent~\cite{zhang2025appagent}, MobileAgent~\cite{wang2025mobile}, and UI-TARS~\cite{qin2025ui}. Deploying PerPilot (powered by the o4-mini model) as a plug-and-play framework, we assess performance using the PerInstruct dataset. PerPilot substantially improves task success rates across all systems: UI-TARS from 12.0\% to 68.0\%, MobileAgent-v2 from 9.3\% to 49.3\%, and AppAgent from 10.7\% to 46.7\%, demonstrating its effectiveness and adaptability. Further, we fine-tune Qwen3-8B~\cite{yang2025qwen3} as \emph{PerQwen}, significantly boosting accuracy from 26.7\% to 74.7\%, closely approaching the closed-source o4-mini model (86.7\%). Additionally, our empirical results indicate that the personalization performance of mobile agents steadily improves with more frequent user interactions. The repository is accessible via the following link: 
https://github.com/xinwang-nwpu/PerPilot

\color{black}

In summary, our contributions are three-fold:
\begin{itemize}
    \item We introduce \emph{PerInstruct}, a novel benchmark composed of personalized instructions spanning diverse daily scenarios, specifically designed to evaluate the personalization capabilities of VLM-based mobile agents.
    \item We propose \emph{PerPilot}, to the best of our knowledge, the first plug-and-play, LLM-powered agent framework capable of effectively perceiving, understanding, and autonomously executing personalized user instructions.
    \item We conduct extensive experiments and detailed analyses, demonstrating that mobile agents equipped with PerPilot achieve state-of-the-art performance on personalized tasks compared to representative existing mobile agent systems.
\end{itemize}

\begin{table*}[ht]
\centering
\fontsize{10pt}{12pt}\selectfont
\setlength{\tabcolsep}{1mm}
\begin{tabular}{lcccccccc}
\hline
 & \makecell{SPA-Bench \\ \shortcite{chen2024spa}} & \makecell{B-MoCA \\ \shortcite{lee2024benchmarking}} & \makecell{MobileAgent \\ Bench \shortcite{wang2024mobileagentbench}} & \makecell{Android \\ World \shortcite{rawles2024androidworld}} & \makecell{AndroidArena \\ \shortcite{AndroidArena}} & \makecell{Android \\ Env \shortcite{toyama2021androidenv}} & \makecell{Mobile-\\ Bench \shortcite{deng2024mobile}} & \textbf{Ours} \\
\hline
\textbf{Real App} & $\surd$ & $\surd$ & $\surd$ & $\surd$ & $\surd$ & $\surd$ & $\surd$ & $\surd$ \\
\textbf{Multi-App} & $\surd$ & $\times$ & $\times$ & $\surd$ & $\surd$ & $\times$ & $\surd$ & $\surd$ \\
\textbf{Number of Apps} & 68 & 16 & 10 & 20 & 15 & 30 & 29 & \textbf{27} \\
\hline
\textbf{Personalization} & $\times$ & $\times$ & $\times$ & $\times$ & $\times$ & $\times$ & $\times$ & $\surd$ \\
\hline
\end{tabular}
\caption{Comparison of \emph{PerInstruct} with existing benchmarks. ``Real App'' indicates the use of real-world Apps; ``Multi-App'' indicates whether tasks span multiple Apps; ``Personalization'' indicates whether personalized tasks are considered. Although PerInstruct contains fewer tasks, our benchmark contains carefully curated, diverse personalized instructions reflecting common real-world mobile scenarios.}

\label{tab:comparison table}
\end{table*}

\section{Related Work}

\subsection{{VLM-based Mobile Agents and Benchmarks}}
Recent studies have proposed various frameworks and benchmarks for VLM-based mobile agents. Frameworks such as AppAgent~\cite{zhang2025appagent}, Mobile-Agent-v2~\cite{wang2025mobile}, and Mobile-Agent-E~\cite{wang2025mobileagente} enable autonomous app exploration, modular workflows, and hierarchical memory for long-term reasoning. Similarly, UI-TARS~\cite{qin2025ui} integrates visual perception and actions via multimodal training and reflective reasoning. To evaluate these agents, benchmarks such as AndroidArena~\cite{xing2024understanding} assess agents' abilities in cross-app interactions, while Mobile-Bench~\cite{deng2024mobile} introduces a hybrid API–UI action space covering single- and multi-app tasks. Other benchmarks, including B-MoCA~\cite{lee2024benchmarking}, AndroidWorld~\cite{rawles2024androidworld}, MobileAgentBench~\cite{wang2024mobileagentbench}, and SPA-BENCH~\cite{chen2024spa}, emphasize agent generalization and robustness across diverse devices and dynamic natural-language tasks.

Despite their diversity and scale, existing frameworks and benchmarks predominantly rely on explicit user instructions, neglecting personalized scenarios where instructions are often ambiguous and involve user-specific preferences. As a result, current VLM-based mobile agents struggle to effectively understand and execute personalized instructions. This highlights the need that specifically enhance and evaluate the personalization capabilities of mobile agents.

\subsection{Personalization in Traditional Mobile Agents}
Personalization---tailoring services to individual preferences~\cite{fan2006personalization,riecken2000introduction}---has drawn significant interest in past few decades. Early approaches primarily filtered content using keyword-based user profiles~\cite{pannu1996learning} or collaborative-filtering techniques based on similar users' preferences~\cite{good1999combining}, later advancing toward adaptive timing and interaction styles~\cite{schiaffino2004user,yorke2009like}. 

However, these methods heavily relied on manual user modeling and extensive domain-specific engineering. Furthermore, existing benchmarks and VLM-based mobile agent systems currently lack consideration of personalization scenarios. Integrating LLM-driven understanding and reasoning capabilities offers a promising pathway to bridging this critical gap.\color{black}

\section{The Famework of PerPliot}
In this section, we present two modules of our \emph{PerPilot}. Due to the lack of space, we flesh out the prompts for the LLM-based operations of this work in Appendix A.

\subsection{Personalization Perception}
Personalization is the process of tailoring content or services to meet the needs of a particular individual, motivated by the recognition that users possess unique requirements~\cite{fan2006personalization,riecken2000introduction}. To effectively fulfill user requests, mobile agents must not only recognize the intent of the instruction but also capture personalized elements---such as characteristics, preferences, interests, and needs---which serve as the foundation for delivering adaptive information and services~\cite{kocaballi2019personalization,akasaki2024detecting}.

Inspired by this, our proposed \emph{PerPilot} firstly perceives whether an instruction needs personalization, which is a prerequisite for using PerPilot. Specifically, the Perception module addresses the question of whether a given user instruction reflects personalized intent and if so, what the relevant personalized element is. We instruct the LLM to identify the elements that may vary across mobile devices or users, or that may even require explicit user input for retrieval. Let $\mathcal{I}$ denote an instruction from a human user (e.g., ``Navigate to my home''), and let $\mathcal{P}$ represent the principles used to assess whether the instruction contains personalized elements. We formalize this process as $\texttt{Perception}(\mathcal{I},\mathcal{P}) \rightarrow (y_{personal}\in\{\texttt{Y,N}\},\{k^i\}_{i=1}^{n})$, where $y_{personal} = \texttt{Y}$ indicates that the instruction $\mathcal{I}$ is personalized. The LLM extracts all corresponding personalized elements $\{k^i\}_{i=1}^{n}$ (e.g., ``my home'' ). Since a single instruction may contain multiple personalized elements, our framework detects all such elements in a single pass. 


\subsection{Personalization Completion}
One of the key characteristics of personalized instructions is that they often contain ambiguous information reflecting the user's preferences~\cite{joko2024doing}. Once an instruction is identified as personalized, it becomes both natural and necessary to complete the missing information related to the personalized elements, enabling the mobile agent to execute the instruction accurately and autonomously. We design the Completion module to fulfill the identified personalized elements through two complementary approaches: memory-based retrieval and reasoning-based exploration.

\subsubsection{Memory-based Retrieval}
For human users, personalized information is typically stable and consistent. For instance, the names of family members rarely change. Based on this characteristic, \emph{PerPilot} constructs a personalized memory database $\mathcal{T}$ that stores user-specific information. This database enables mobile agents to execute personalized requests more efficiently and autonomously in future interactions. Initially, the database $\mathcal{T}$ is empty. However, as the number of interactions increases, PerPilot progressively accumulates and refines its understanding of each user's personalized information. Over time, this allows the mobile agent to execute requests more accurately and to become increasingly intelligent through continual use.

Specifically, when an instruction $\mathcal{I}$ is received and identified as personalized, we instruct the LLM to first retrieve the relevant personalized information from the database $\mathcal{T}$. If the required information is found, the LLM completes the instruction and outputs `Y' and $\mathcal{I}_{completed}$ accordingly. If not, the LLM returns `N' with the personalized elements $\{k^i\}_{i=1}^{n}$. We formalize this memory-based retrieval as: 
\begin{multline*}
    \texttt{RetrievalCompletion}(\mathcal{I}, \mathcal{T}, \{k^i\}_{i=1}^{n}) \rightarrow \\
    \begin{cases}
    (\texttt{Y},\mathcal{I}_{completed}), & \text{if } \forall k^i, \ k^i \in \mathcal{T} \\
    (\texttt{Part}, \mathcal{I}_{part}, \{k^i\}_{i=1}^{m}), & \text{if } \exists k^i \notin \mathcal{T} \text{ and } \exists k^j \in \mathcal{T} \\
    (\texttt{N}, \{k^i\}_{i=1}^{n}). & \text{if } \forall k^i, \ k^i \notin \mathcal{T}
    \end{cases}
\end{multline*}
Here, $\mathcal{I}_{part}$ denotes the instruction that has been partially completed, while $\{k^i\}_{i=1}^{m}$ represents the remaining personalized elements that require further exploration. Due to space constraints, illustrative examples demonstrating how the framework operates are provided in Appendix B.



\subsubsection{Reasoning-based Exploration}
If personalized information is unavailable in the database $\mathcal{T}$, the simplest solution is to prompt the user to explicitly clarify the instruction, enabling direct execution. However, frequently prompting users negatively impacts convenience and user experience. We observe that such personalized information typically exists in widely-used mobile applications---for instance, addresses (e.g., ``home'') in shopping apps, or contacts (e.g., ``mom'') in social apps. To address this, the \emph{Exploration} component of PerPilot autonomously generates a specific instruction, guiding the mobile agent to locate the missing personalized information within a relevant app. Leveraging the reasoning capabilities of LLMs, our framework infers the most likely app containing the required data and formulates an exploration instruction accordingly. Formally, given an instruction $\mathcal{I}$ (e.g., ``Navigate to my home''), unresolved personalized elements $\{k^i\}_{i=1}^{m}$ (e.g., ``my home''), and a list of available apps $\mathcal{L}_{app}$ on the device, this reasoning-based exploration process is defined as $\texttt{ExploreCompletion}(\mathcal{I}, \{k^i\}_{i=1}^{m}, \mathcal{L}_{app}) \rightarrow \mathcal{I}_{app}$, where $\mathcal{I}_{app}$ directs the mobile agent to retrieve the missing personalized information (e.g., ``From the app Taobao, retrieve my home address''). Illustrative examples demonstrating this process are provided in Appendix~B due to space constraints.

~\\
Once PerPilot retrieves the required personalized information, it will complete the user's instruction accordingly and then execute it autonomously. If PerPilot cannot retrieve the information, it will prompt the user to clarify the instruction.

\section{PerInstruct}
In this section, we introduce \emph{PerInstruct}, the first human-annotated dataset containing personalized instructions across diverse mobile scenarios. We describe the dataset construction and provide a detailed evaluation analysis.
\subsection{Data Construction}

\textbf{Basic Information of \emph{PerInstruct}.} Our benchmark comprises $75$ carefully curated instructions referencing $27$ widely-used mobile apps. These instructions incorporate $28$ personalized elements---terms whose meanings vary among users (e.g., ``mom'', ``home'', ``favorite song'')---all selected based on the functionalities of these apps. For clarity, we present a representative yet straightforward instruction in the box below. The complete dataset is provided in Appendix C.
\begin{tcolorbox}[title=An Example in \textit{PerInstruct}.]
\textbf{Instruction ID:} 20.\\
\textbf{Natural-language instruction:} Open QQ to view the questions sent by my friend and copy it.\\
\textbf{Difficulty label:} simple.\\
\textbf{The minimal step count:} 4.\\
\textbf{The app(s) involved:} QQ.\\
\textbf{The completed form of instruction:} Open QQ to view the questions sent by \{Name\} and copy it.\\
\textbf{Personalized element:} my friend.\\
\textbf{Personalized info:} \{Name\}.
\label{box}
\end{tcolorbox}

\begin{figure}[ht]
  \centering
  \includegraphics[width=0.75\linewidth]{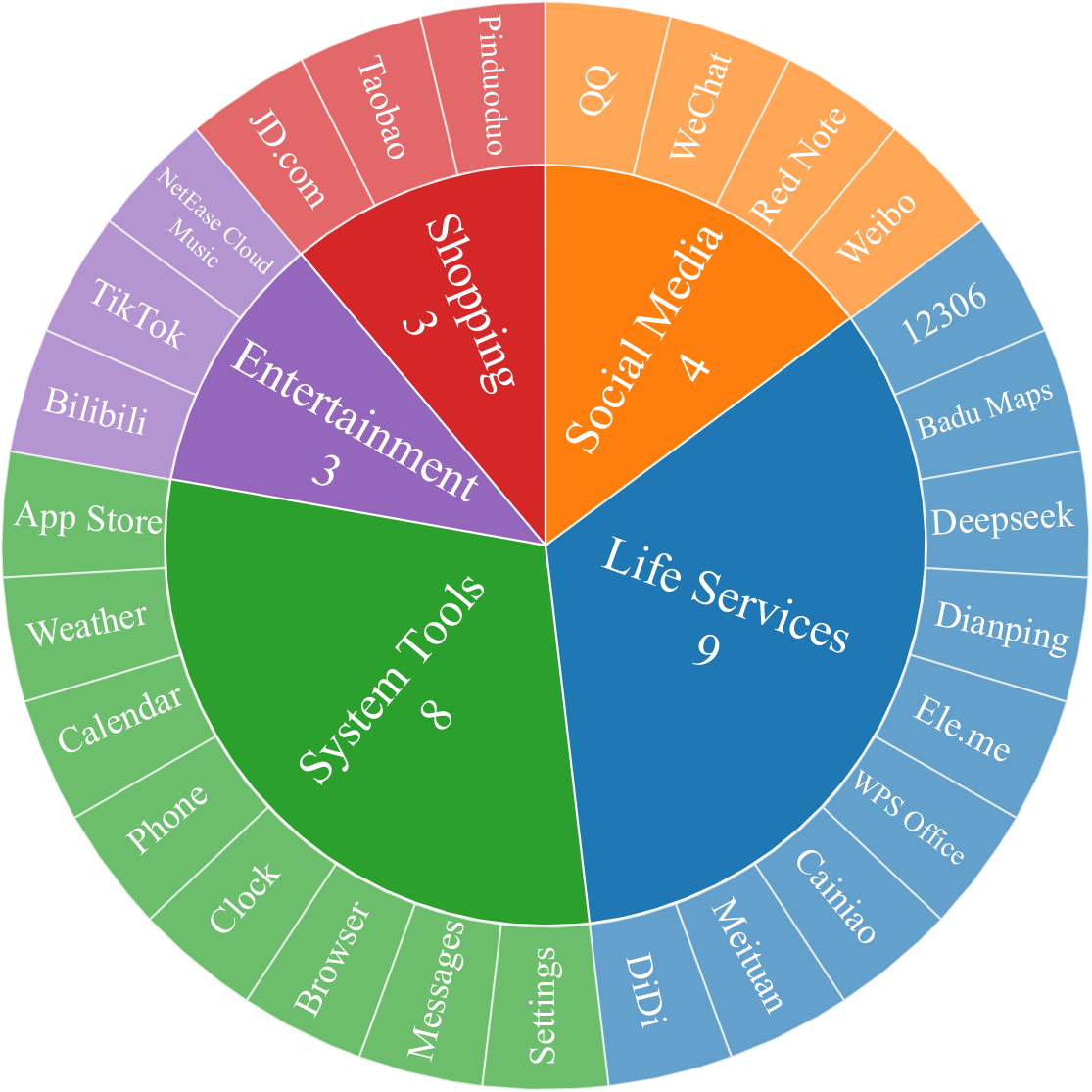}
  \caption{Categories of Apps covered by \emph{PerInstruct}}
  \label{fig:APP category}
\end{figure}

\textbf{App selection and categorization.} As shown in Figure \ref{fig:APP category}, we select $27$ popular mobile apps, categorized into five groups: (i) \emph{Social and Communication} for messaging and calls; (ii) \emph{System Tools} for basic device functions; (iii) \emph{Life Services} supporting daily tasks like payments and navigation; (iv) \emph{Entertainment} providing leisure content; and (v) \emph{Shopping} enabling mobile commerce. Collectively, these apps span the dominant activities of everyday mobile use.

\textbf{Difficulty categorization.} To effectively assess varying levels of personalization complexity, instructions are categorized into three types based on the number of personalized elements and the apps involved: (i) \emph{simple}: instructions contain a single personalized element and involve a straightforward, single-step interaction within one app (e.g., ``Open Alipay and pay my rent''); (ii) \emph{normal}: instructions incorporate multiple personalized elements and require several steps but are confined to a single application; (iii) \emph{hard}: instructions integrate multiple personalized elements into complex, multi-step workflows spanning multiple applications.

\subsection{Data Evaluation}

 We evaluated \emph{PerInstruct} using both quantitative and qualitative evaluation. First, we propose two metrics:
\begin{itemize}
    \item \textit{Difficulty Label Consistency (DLC).} We measure alignment between assigned difficulty labels (i.e., simple, normal, hard) and actual execution difficulty (i.e., step count) using Spearman correlation; a correlation $\ge0.70$ indicates valid labeling.

    \item \textit{Distribution Entropy (DE).} We measure how uniformly tasks are distributed across dimensions such as difficulty level and app diversity. It quantifies whether instructions are evenly spread or disproportionately concentrated in certain categories. A value closer to $1$ indicates a perfectly balanced distribution across categories, while a value approaching $0$ signals significant imbalance or skew.
\end{itemize}

\begin{table}[!ht]
  \centering
  \fontsize{10pt}{12pt}\selectfont
  \label{tab:quality}
  \begin{tabular}{c|c|c}
    \toprule
    \textbf{Category} & \textbf{Metric} & \textbf{ Value} \\
    \midrule
    \multirow{3}{*}{Quantitative Metrics}
      & DLC & $0.86$ \\
      & $\text{DE}_{\text{difficulty}}$ & $0.98$ \\
      & $\text{DE}_{\text{diversity}}$ & $0.89$ \\
    \midrule
    \multirow{5}{*}{Human Evaluation}
      & personalization & $4.44 \pm 0.32$\\
      & difficulty & $4.40 \pm 0.32$\\
      & accuracy & $4.69 \pm 0.19$\\
      & coverage & $4.00	\pm 0.00$\\
      & diversity & $4.33	\pm 0.51$\\
    \bottomrule
  \end{tabular}
  \caption{Evaluation for \emph{PerInstruct} from quantitative metrics and human experts. ``DLC'' measures rationality of difficulty categories; ``DE'' evaluates balanced distribution of tasks. For human evaluation, the average scores is $4.37 \pm 0.26$ on a 5-point Likert scale (1: strongly disagree, 5: strongly agree).}
  \label{tab: dataset quality}
\end{table}

Second, we recruited seven human experts to provide a gold-standard evaluation. Each expert independently scored every instruction on five aspects using five-point Likert scales: (i) personalization: whether each instruction contains personalized elements; (ii) difficulty: whether the assigned difficulty level is appropriate; (iii) accuracy: whether the extracted personalized elements are correct and complete; (iv) coverage: whether the instructions collectively capture most daily mobile-usage scenarios; and (v) diversity: whether the dataset includes a wide variety of apps. Full evaluation details are provided in Appendix C.

Table \ref{tab: dataset quality} summarizes results from both quantitative analyses and human evaluations. For difficulty label consistency, \emph{PerInstruct} achieves a score of $0.86$, indicating strong consistency between labeled difficulty and execution step count. Additionally, the distribution entropy scores indicate a well-balanced dataset, with values of $0.98$ for difficulty levels and $0.89$ for app diversity. Furthermore, human evaluators provided positive feedback on our dataset, with an overall average rating of $4.37 \pm 0.26$, and ratings above $4$ across all five aspects (with $3$ indicates ``neutral'' and $5$ ``strongly agree'').
\color{black}




\begin{table*}[!t]
\centering
\fontsize{10pt}{12pt}\selectfont
\setlength{\tabcolsep}{1mm}
\begin{tabular}{llcccccc}
\toprule
\multirow{2}{*}{\makecell{\raggedright \textbf{Base} \textbf{System}}} & \multirow{2}{*}{\makecell{Base Model of PerPilot}} & \multirow{2}{*}{\makecell{Human Intervention \\ Count}} & \multicolumn{4}{c}{\textbf{SuccessRate}} & \multirow{2}{*}{\makecell{\textbf{vs.} \\ \textbf{Baseline}}} \\
\cmidrule(lr){4-7}
& & & Simple & Normal & Hard & Overall & \\
\midrule
\multirow{5}{*}{\makecell[l]{UI-TARS \\ \cite{qin2025ui}}} & without PerPilot & - & 28.1\% & 0\% & 0\% & 12.0\% (9/75) & - \\
\cmidrule(lr){2-8}
& PerQwen & - & 65.6\% & 38.1\% & 40.9\% & 50.7\% (38/75) & \textcolor{black}{+38.7} \\
& o4-mini & - & 65.6\% & 61.9\% & 36.4\% & 56.0\% (42/75) & \textcolor{black}{+44.0} \\
& PerQwen${}_{\text{(with Human Intervention)}}$  & 7 & 71.9\% & 71.4\% & 40.9\% & 62.7\% (47/75) & \textcolor{black}{+50.7} \\
& o4-mini${}_{\text{(with Human Intervention)}}$  & 6 & \textbf{81.3\%} & \textbf{81.0\%} & \textbf{36.4\%} & \textbf{68.0\% (51/75)} & \textcolor{black}{\textbf{+56.0}} \\  \midrule

\multirow{5}{*}{\makecell[l]{MobileAgent-v2 \\ \cite{wang2025mobile}}} & without PerPilot & - & 21.9\% & 0\% & 0\% & 9.3\% (7/75) & - \\
\cmidrule(lr){2-8}
& PerQwen & - & 34.4\% & 9.5\% & 9.1\% & 20.0\% (15/75) & \textcolor{black}{+10.7} \\
& o4-mini  & - & 43.8\% & 9.5\% & 9.1\% & 24.0\% (18/75) & \textcolor{black}{+14.7} \\
& PerQwen ${}_{\text{(with Human Intervention)}}$  & 8 & 62.5\% & 28.6\% & 22.7\% & 41.3\% (31/75) & \textcolor{black}{+32.0} \\
& o4-mini${}_{\text{(with Human Intervention)}}$ & 6 & 68.8\% & 38.1\% & 31.8\% & 49.3\% (37/75) & \textcolor{black}{+40.0} \\  \midrule

\multirow{5}{*}{\makecell[l]{AppAgent \\ \cite{zhang2025appagent}}} & without PerPilot & - & 25.0\% & 0\% & 0\% & 10.7\% (8/75)& - \\
\cmidrule(lr){2-8}
& PerQwen & - & 28.1\% & 4.8\% & 9.1\% & 16.0\% (12/75) & \textcolor{black}{+5.3} \\
& o4-mini & - & 40.6\% & 4.8\% & 4.5\% & 20.0\% (15/75) & \textcolor{black}{+8.0} \\
& PerQwen${}_{\text{(with Human Intervention)}}$  & 12 & 56.3\% & 42.9\% & 9.1\% & 38.7\% (29/75) & \textcolor{black}{+28.0} \\
& o4-mini${}_{\text{(with Human Intervention)}}$ & 12 & 68.8\% & 57.1\% & 4.5\% & 46.7\% (35/75) & \textcolor{black}{+34.7} \\
\bottomrule
\end{tabular}%
\caption{Overall performance comparison of mobile agents equipped with PerPilot across three representative agent systems. We evaluate the baseline agents (without PerPilot) and agents enhanced by PerPilot powered by either the o4-mini model or our fine-tuned model, PerQwen. The notation ``with Human Intervention'' indicates settings involving human intervention. The SuccessRate column reports success rates across simple, normal, hard, and overall datasets. The last column shows absolute improvement in SuccessRate over baseline performance. Best results are highlighted in bold.}
\label{tab:o4-mini-performance}
\end{table*}

\section{Experiment}

Our experiment aims to answer three questions: (i) Can PerPilot effectively enhance the personalization capability of mobile agents? (ii) Can PerPilot serve as a plug-and-play framework that adapts to diverse mobile agent systems? (iii) If so, how well does our framework perform on our benchmark? We outline the experimental settings and then answer the corresponding questions in this section.

\subsection{Experimental Settings}
\label{sec:setup}

\textbf{Baselines and Our \textit{PerPilot}.} We select three representative mobile-agent systems as baselines: AppAgent~\cite{zhang2025appagent} and MobileAgent-v2~\cite{wang2025mobile}, both powered by the o4-mini model, and UI-TARS~\cite{qin2025ui}, which directly models agents. To comprehensively assess the effectiveness and robustness of our approach, we integrate \emph{PerPilot}, powered by the o4-mini-2025-04-16 model~\cite{openai2025o3o4mini}, as a plug-and-play personalization framework into each baseline agent. Additional experimental details are provided in Appendix D. Experimental code is available at: https://github.com/xinwang-nwpu/PerPilot

\textbf{PerQwen.} To reduce reliance on closed-source models and address potential privacy issues intrinsic to personalized tasks, we further present \emph{PerQwen}, a fine-tuned variant of the open-source Qwen3-8B model~\cite{yang2025qwen3}, specifically optimized for personalized agent capabilities. \color{black}

\textbf{Evaluation Metrics.} To evaluate whether PerPilot effectively enables mobile agents to perceive, understand, and execute personalized user instructions, we consider four metrics: (i) \textit{Success rate}\color{black}, measuring overall performance as the proportion of instructions successfully completed by the agent; (ii) \textit{Element perception accuracy}\color{black}, assessing the agent's capability to identify personalized elements, quantified as the proportion of instructions from which personalized element is correctly recognized; (iii) \textit{Exploration accuracy}\color{black}, evaluating the agent’s capability to explore and retrieve relevant personalized information, quantified by the proportion of instructions for which such information is successfully obtained; and (iv) \textit{Human intervention count}\color{black}, quantifying how frequently the agent requests human intervention, thus reflecting its dependence on human intervention.

\subsection{Experimental Results and Analysis}
\label{sec:results}

\textbf{\emph{PerPilot} effectively enhances the personalization capability of mobile agents.} Table~\ref{tab:o4-mini-performance} summarizes agents’ performance with PerPilot across tasks of varying difficulty. Notably, the UI-TARS-based agent improved significantly from a baseline success rate of 12.0\% to 62.7\% (PerQwen-driven) and 68.0\% (o4-mini-driven), representing absolute increases of 50.7 and 56.0 points, respectively. These results highlight PerPilot’s effectiveness in enhancing personalized instruction execution with minimal human intervention.
%

\begin{table}[h!]
\centering
\fontsize{10pt}{12pt}\selectfont
\begin{tabular}{lccc}
\toprule
\textbf{Metric} & \textbf{Qwen3-8B} & \textbf{PerQwen} & \textbf{o4-mini} \\ 
\midrule
EP Acc. & 26.7\% & 74.7\% & 86.7\% \\
Ex Acc.${}_{w/o~HI}$ & 22.7\% &  60.0\% & 73.3\% \\
Ex Acc.${}_{with~HI}$ & 26.7\% & 74.7\% & 86.7\% \\
HI Count & 3 & 7 & 6 \\ \hline
\textbf{SuccessRate} & 20.0\% & 62.7\% & 68.0\%  \\ 
\bottomrule
\end{tabular}
\caption{Performance comparison of PerPilot across models (Qwen3-8B, PerQwen, o4-mini). \textit{Element perception accuracy (EP Acc.)} indicates accuracy in recognizing personalized elements; \textit{Exploration accuracy (Ex Acc.)} measures the success rate in retrieving personalized information (\textit{with HI}: with human intervention); \textit{Human intervention count (HI Count)} denotes the total number of user assistance for personalized information.}
\label{tab:personalized_results}
\end{table}

\begin{figure*}[!t]
  \centering
  \includegraphics[width=0.97\linewidth]{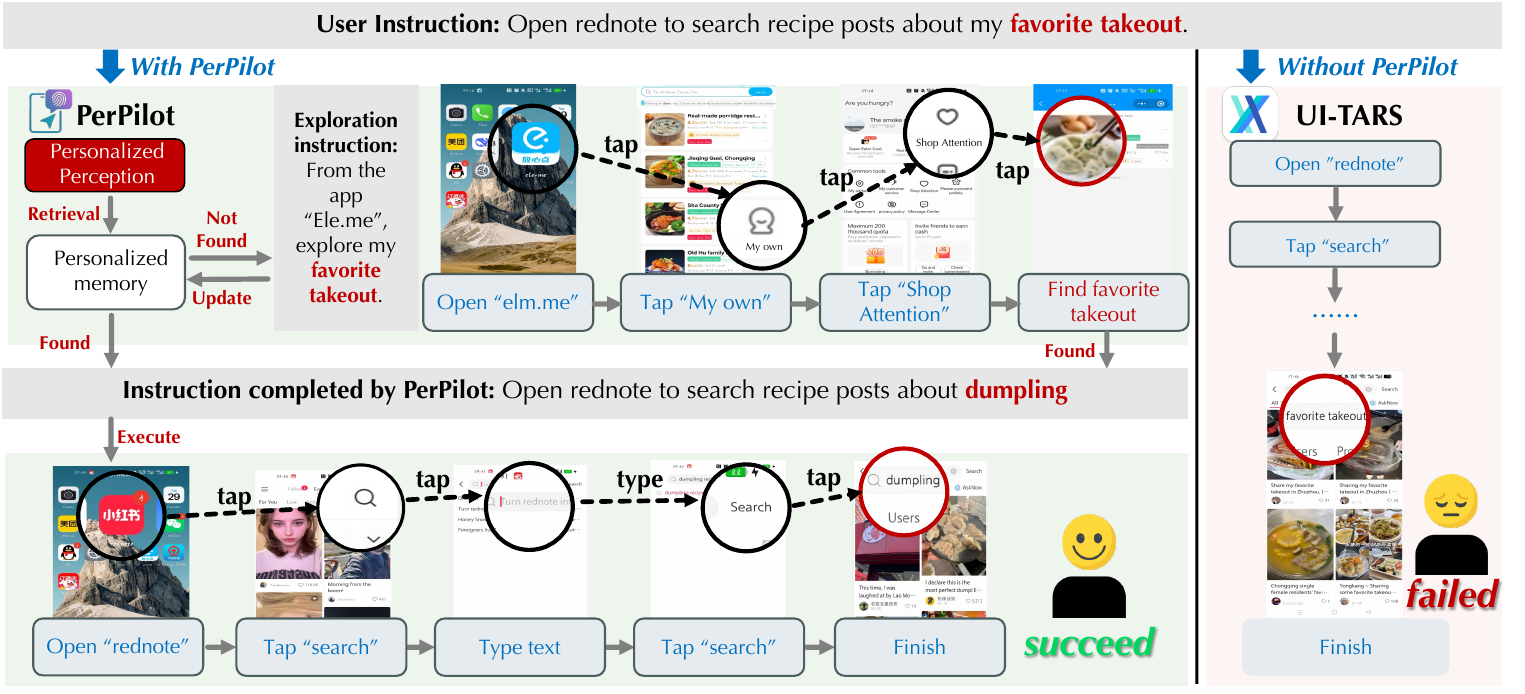}
  \caption{A case study illustrating how a personalized instruction (searching for recipes of ``my favorite takeout'') is successfully perceived, understood, and executed by mobile agents equipped with PerPilot, in contrast to the failure of a baseline agent system without PerPilot.}
  \label{fig:case}
\end{figure*}

\textbf{\emph{PerPilot} serves as a robust and effective plug-and-play framework adaptable to various mobile agent systems.} To validate the robustness of PerPilot, we deployed it across three different mobile agent systems---UI-TARS, MobileAgent-v2, and AppAgent---as presented in Table~\ref{tab:o4-mini-performance}. Each system exhibited notable improvements in overall success rates upon integration with PerPilot. Specifically, the UI-TARS system showed an increase from 12.0\% to 68.0\%, the MobileAgent-v2 system increased from 9.3\% to 49.3\%, and the AppAgent system improved from 10.7\% to 46.7\%. The consistent improvements in success rates across diverse agent systems clearly demonstrate PerPilot's adaptability and effectiveness in enhancing the personalization capabilities of various mobile agents.


\textbf{PerPilot, powered by PerQwen, achieves performance comparable to the o4-mini-driven version in perceiving personalized instructions and exploring relevant information.} Table~\ref{tab:personalized_results} quantitatively illustrates these results. Specifically, PerQwen notably outperforms baseline Qwen3-8B in element perception accuracy (74.7\% vs.\ 26.7\%), closely approaching o4-mini's 86.7\%. Additionally, PerQwen significantly improves information exploration accuracy from 22.7\% to 60.0\% (74.7\% with minimal human assistance) and increases overall task success from 20.0\% to 62.7\%,closely approaching the 68.0\% achieved by the o4-mini-driven version.

\begin{figure}[ht]
  \centering
  \includegraphics[width=0.9\linewidth]{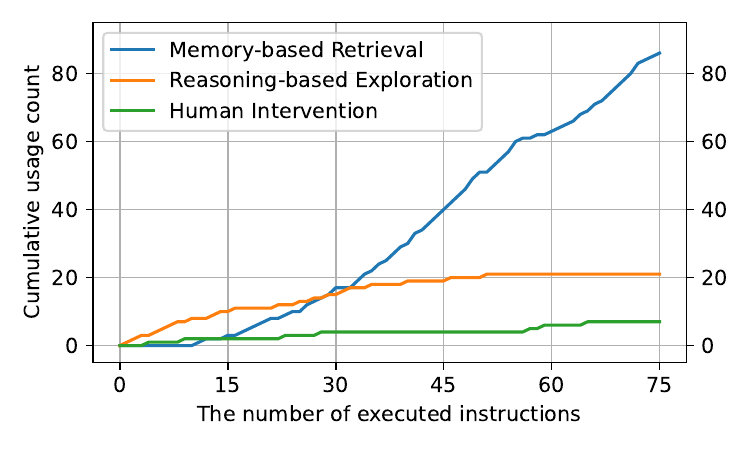}
  \caption{Frequency of three approaches for obtaining personalized information. As the number of executed instructions increases, agents increasingly rely on memory-based retrieval rather than reasoning-based exploration or human intervention, reflecting more personalization over time.}
  \label{fig:explore}
\end{figure}

\textbf{The more frequently users instruct our agent, the smarter and more personalized it becomes.} As shown in Figure~\ref{fig:explore}, PerPilot initially relies on exploration to gather personalized information, occasionally requiring human intervention. With continued use, PerPilot increasingly retrieves information directly from memory, reducing both exploration and human assistance, thus improving its responsiveness and personalization capabilities.

\color{black}

\subsection{A Case Study}
Figure~\ref{fig:case} presents a case study illustrating how PerPilot perceives, understands, and executes a personalized user instruction. Specifically, the user instructs the agent to search for recipes related to ``my favorite takeout''---a personalized element since preferences vary among individuals and are likely stored within user mobile apps. PerPilot first identifies personalized elements and attempts to retrieve corresponding information from memory. If unavailable, PerPilot autonomously reasons about relevant apps (e.g., lifestyle service apps like ``Ele.me''), generates an exploration instruction, and interacts with the mobile interface through human-like operations to obtain the missing information (e.g., ``dumpling''). Once clarified, PerPilot completes the original instruction by opening the ``Rednote'' app and successfully searching for dumpling recipes. In contrast, a baseline agent without PerPilot directly searches for ``my favorite takeout,'' resulting in task failure due to missing personalized information.\color{black}

\section{Conclusion}
The study of mobile agents has traditionally focused on assisting users through human-like interactions based on explicit instructions. However, personalized instructions---those reflecting individual preferences or requirements---have largely been overlooked. To this end, we define personalized instructions and introduce a novel benchmark covering a wide range of personalized daily scenarios. Furthermore, to enhance mobile agents’ capabilities in perceiving, understanding, and executing personalized instructions, we propose \textit{PerPilot}, the first LLM-powered mobile agent framework designed as a plug-and-play module compatible with existing mobile agent systems. PerPilot enables agents to accurately identify personalized elements within instructions and autonomously complete these instructions via two complementary approaches: memory-based retrieval and reasoning-based exploration. We evaluated PerPilot using both a closed-source model (o4-mini) and our fine-tuned model (PerQwen) integrated with three leading mobile agent systems. Extensive experimental results demonstrate that PerPilot significantly enhances the personalization capabilities of mobile agents. Additionally, we empirically demonstrate that the personalization capability of PerPilot improves continuously with more frequent user instructions.



\bigskip

\bibliography{aaai2026}

\newpage
\appendix
\onecolumn
\section{Appendix A\hspace{1em}Prompt Details} 
Prompts used in PerPilot can be divided into two categories. The first category comprises prompts that guide the large language model (LLM) within the PerPilot framework, applied during the stages of Personalization Perception and Personalization Completion (i.e., reasoning-based exploration). The second category consists of prompts guiding the large multimodal model (VLM) employed by agents integrated with PerPilot, assisting them during the stage of executing instructions.
\subsection*{A.1 PerPilot Prompt}
\textbf{Prompt for \textit{Personalization Perception}.}
This prompt empowers the agent to first identify personalized elements within input instructions. In this process, we prompt the LLM to perceive personalized elements as those exhibiting varying preferences among individual users or those explicitly requiring user input for retrieval. Given that a single instruction can encompass multiple personalized elements, our framework accomplishes the detection of all such elements in a single pass.

\begin{tcolorbox}[title={Prompt for Personalization Perception}]

\textbf{INPUT:} Please understand and evaluate the instructions I have given you to determine if they contain personalized elements.  
If the instruction contains words that need to be clarified by asking the user, or if certain words have different meanings for different people or devices, it can be determined that the instruction contains personalized elements, and these words are personalized elements.

You need to strictly follow the following rules:  

Rule 1: If certain words have unique executable meanings, such as app names like QQ and WeChat, they are not personalized elements.  

Rule 2: When you think something is not a personalized element, directly determine that it is not a personalized element.  

Rule 3: Strictly prohibit treating specific names as personalized elements, whether they are Chinese or English names. But abstract names are still personalized elements, such as friends.  

\vspace{4pt} 

\textbf{OUTPUT:}  
If you think it is not a personalized instruction, please answer `No'.  

If you think this is a personalized instruction, you need to determine which part of the instruction is the personalized element.  

Then your answer should follow this format:  
`Yes\textbar First personalized element (i.e., the first part you consider personalized)\textbar Second personalized element\textbar Third personalized element (and so on, output all personalized elements,The same element only needs to be output once)'  
The current instruction is as follows:\{instruction\}

Please note that your answer should not include any additional information outside the format provided.  

\end{tcolorbox}


\noindent\textbf{Prompt for Information Exploration.}
This prompt empowers the agent to leverage the characteristic that personalized information typically exists in widely used mobile apps. It further utilizes the reasoning capabilities of the LLM to infer the most probable application containing the required information, subsequently formulating an appropriate exploration instruction. This approach facilitates the accurate retrieval of missing personalized information.

\begin{tcolorbox}[title={Prompt for Information Exploration}]
\textbf{INPUT:} You need to assist me in completing an information exploration task. The specific task information is as follows:  
You are currently controlling the user's phone to complete the personalized instruction \{instruction\}, but those personalized information: \{search\_element\} (You only need to deal with the personalized elements in parentheses) in the instruction is missing. 
I am now trying to obtain the precise information for these personalized elements from the user's phone.  

It is known that the user's phone has the following apps:
\{app\_lists\} 

Please carefully consider the types of these apps and the information they may contain. For each personalized element, select the app that is most likely to store the corresponding precise information (note that each personalized element can only select one app, do not select multiple apps).  

\vspace{4pt}

\textbf{OUTPUT:} Output the same number of instructions as the number of personalized elements (do not output extra instructions, only output one instruction per personalized element, strictly forbidden to output extra instructions), each instruction should be in the format 'From the app XX, obtain the YY (the YY is personalized element and personalized element must be included in the sentence) XX information (here, XX is the type of information you need to obtain, for example From QQ, obtain the friend name information. please note one instruction per line)'.  
\end{tcolorbox}

\subsection*{A.2 Guidance Prompt for Agent Execution}

This prompt provides additional guidance to assist the agent in accurately and effectively identifying personalized information during exploratory interactions. It instructs the agent to correctly interpret personalized terms, avoid ambiguous searches, and prioritize accuracy and relevance based on user intent.

\begin{tcolorbox}[title={Guidance Prompt for Appagent}]
\textbf{INPUT:} You are an agent that is trained to perform some basic tasks on a smartphone. You will be given a 
smartphone screenshot. The interactive UI elements on the screenshot are labeled with numeric tags starting from 1. The 
numeric tag of each interactive element is located in the center of the element.

 You need to help the user find the corresponding information in this app based on their instructions. The following hints may help you better complete the task.

    Hint 1: The information the user needs contains personalized elements, which have different meanings for different people, such as home, friends. Therefore, do not directly search for these terms.  

    Hint 2: When you find information marked with ... (ellipsis), you should try to obtain the full content of the information rather than directly outputting the information with ellipsis.  

    Hint 3: The user's original instruction is \{instruction\}. The information in this instruction may help you better find the information the user needs.  

    Remember that the above hints are only auxiliary information; you need to use your own judgment to determine if they are useful.  

    You need to use your thinking ability to first determine what information to find.  

    Note that for the task, more information is not necessarily better; rather, more concise information is better (for example, for home, you usually only need to find an address; for a good friend, you usually only need to find a name. For other types of information, think about what you need to find). Then find which information in this app is most likely to represent the information you need

\textbf{OUTPUT:}
     If you believe the task is completed or there is nothing to be done, you should output FINISH\textbar information (this information is the core part of what the user needs)).

\end{tcolorbox}

\begin{tcolorbox}[title={Guidance Prompt for MobileAgent-v2}]
\textbf{INPUT:} This image is a phone screenshot. Its width is \{width\} pixels and its height is \{height\} pixels. The user\'s instruction is: \{instruction\}.

 You need to help the user find the corresponding information in this app based on their instructions. The following hints may help you better complete the task.

    Hint 1: The information the user needs contains personalized elements, which have different meanings for different people, such as home, friends. Therefore, do not directly search for these terms.  

    Hint 2: When you find information marked with ... (ellipsis), you should try to obtain the full content of the information rather than directly outputting the information with ellipsis.  

    Hint 3: The user's original instruction is \{instruction\}. The information in this instruction may help you better find the information the user needs.  

    Remember that the above hints are only auxiliary information; you need to use your own judgment to determine if they are useful.  

    You need to use your thinking ability to first determine what information to find.  

    Note that for the task, more information is not necessarily better; rather, more concise information is better (for example, for home, you usually only need to find an address; for a good friend, you usually only need to find a name. For other types of information, think about what you need to find). Then find which information in this app is most likely to represent the information you need

\textbf{OUTPUT:}
     If you believe all the requirements of the user's instruction have been completed and no further action is needed, you can choose this operation to terminate the process.Then your output format is Stop\textbar Information (this information is the core part of what the user needs)

\end{tcolorbox}

\begin{tcolorbox}[title={Guidance Prompt for UI-TARS}]
\textbf{INPUT:} You are assisting a user with a mobile command task.The user's instruction is: \{instruction\}.

 You need to help the user find the corresponding information in this app based on their instructions. The following hints may help you better complete the task.  

    Hint 1: The information the user needs contains personalized elements, which have different meanings for different people, such as home, friends. Therefore, do not directly search for these terms.  

    Hint 2: When you find information marked with ... (ellipsis), you should try to obtain the full content of the information rather than directly outputting the information with ellipsis.  

    Hint 3: The user's original instruction is \{instruction\}. The information in this instruction may help you better find the information the user needs.  

    Remember that the above hints are only auxiliary information; you need to use your own judgment to determine if they are useful.  

    You need to use your thinking ability to first determine what information to find.  

    Note that for the task, more information is not necessarily better; rather, more concise information is better (for example, for home, you usually only need to find an address; for a good friend, you usually only need to find a name. For other types of information, think about what you need to find). Then find which information in this app is most likely to represent the information you need

\textbf{OUTPUT:}
     If you believe all the requirements of the user's instruction have been completed and no further action is needed, you can choose this operation to terminate the process.Then your output format is Stop\textbar Information (this information is the core part of what the user needs)

\end{tcolorbox}

\newpage
\section{\textbf{Appendix B}\hspace{1em}\textbf{PerPliot's Model Example}} 
We present examples demonstrating how the PerPilot framework operates, including \textit{Personalization Perception} and \textit{Personalization Completion (i.e., memory-based retrieval and reasoning-based exploration)}.
\subsection{B.1 Personalization Perception Example}

\begin{tcolorbox}[title={Example of Personalization Perception}]

\textbf{INPUT:}
Please understand and evaluate the instructions I have given you to determine if they contain personalized elements.If the instruction contains words......

Rule 1: If certain words have unique executable meanings, such as app names......

Instruction: Open TikTok, search for a video about my school, and share it with my friend.

Desired format: If you think it is not a personalized instruction, please answer ‘No’.If you think this is a personalized instruction, Then your answer should follow this format: `Yes\textbar First personalized element\textbar Second personalized element......’

\vspace{4pt}

\textbf{OUTPUT:} Yes\textbar my school\textbar my friend  

\end{tcolorbox}

\begin{tcolorbox}[title={Example of Personalization Perception}]

\textbf{INPUT:}
Please understand and evaluate the instructions I have given you to determine if they contain personalized elements.If the instruction contains words......

Rule 1: If certain words have unique executable meanings, such as app names......

Instruction: Open TikTok, search for a video about Mr.Beast, and share it with Bob.

Desired format: If you think it is not a personalized instruction, please answer ‘No’.If you think this is a personalized instruction, Then your answer should follow this format: `Yes\textbar First personalized element\textbar Second personalized element......’

\vspace{4pt}

\textbf{OUTPUT:} No

\end{tcolorbox}
\newpage
\subsection{B.2 Personalization Completion Example}

\begin{tcolorbox}[title={Example of Memory-based Retrieval}]
\textbf{INPUT:}Instruction: Open TikTok, search for a video about my school, and share it with my friend. 

Personalized elements: my school\textbar my friend

Memory: my school\textbar test school......

\textbf{OUTPUT:} No\textbar my friend

\end{tcolorbox}

\begin{tcolorbox}[title={Example of Reasoning-based Exploration}]

\textbf{INPUT:}
You are currently controlling the user’s phone to complete the personalized instruction \{instruction\}, but those personalized information: \{search element\}in the instruction is missing. I am now trying to obtain the precise information for these personalized elements......

It is known that the user’s phone has the following apps: \{app lists\} Please carefully consider the types of these apps and the information they may contain......

Desired format: Output the same number of instructions as the number of personalized elements. Each instruction should
be in the format ’From the app XX, obtain the YY information......

Instruction: Open TikTok, search for a video about my school, and share it with my friend.  

search\_element: [`my friend'] 

\textbf{OUTPUT:} From the app wechat, obtain my friend name information.

\end{tcolorbox}

\begin{tcolorbox}[title={Example of Agent Exploration}]
\textbf{INPUT:}
You need to help the user find the corresponding information in this app based on their instructions......

Hint 1: The information the user needs contains personalized elements, which have different meanings for different people......

Desired format:If you believe all the requirements of the user’s instruction have been completed and no further action is needed,
you can choose this operation to terminate the process.Then your output format is Stop\textbar Information......

Instruction: From the app wechat, obtain my friend name information.

\textbf{OUTPUT:} Stop\textbar jack

\end{tcolorbox}

\begin{tcolorbox}[title={Example of Memory-based Retrieval}]
\textbf{INPUT:} Instruction: Open TikTok, search for a video about my school, and share it with my friend. 

Personalized elements: my school\textbar my friend

Memory: my school\textbar test school, my friend\textbar jack......

\textbf{OUTPUT:} Open TikTok, search for a video about test school, and share it with jack. 

\end{tcolorbox}

\newpage

\section{\textbf{Appendix C}\hspace{1em}\textbf{PerInstruct's details}}

The following section presents the complete PerInstruct dataset and describes the methodology used by human experts for evaluation.

\subsection*{C.1 Complete \textit{PerInstruct} Dataset}

\tcbset{
  outer-box/.style={
    breakable=true, 
    boxrule=\fboxrule, 
    top=0pt,
    bottom=0pt,
    left=0pt,
    right=0pt,
    boxsep=\fboxsep, 
    colframe=black, 
    colback=white, 
    title style={font=\bfseries}, 
    before skip=\baselineskip, 
    after skip=\baselineskip, 
  },
  instruction-block/.style={
    nobreak=true, 
    colback=white,
    colframe=white, 
    boxrule=0pt,
    left=0pt, right=0pt, top=0pt, bottom=0pt,
    boxsep=0pt, 
    before skip=0pt,
    after skip=0pt
  }
}

\begin{tcolorbox}[outer-box, title={Complete Dataset}]

\textbf{Instruction ID:} 1.\\
\textbf{Natural-language instruction:} Open Map and navigate to xian.\\
\textbf{Difficulty label:} easy.\\
\textbf{The minimal step count:} 6.\\
\textbf{The app(s) involved:} Baidumap.\\
\textbf{The completed form of instruction:} Open Map and navigate to \{place name\}.\\
\textbf{Personalized element:} .\\
\textbf{Personalized info:} place name.
\label{box1}

\vspace{4pt}
\hrulefill
\vspace{4pt}

\textbf{Instruction ID:} 2.\\
\textbf{Natural-language instruction:} call David.\\
\textbf{Difficulty label:} easy.\\
\textbf{The minimal step count:} 4.\\
\textbf{The app(s) involved:} Phone.\\
\textbf{The completed form of instruction:} call \{name\}.\\
\textbf{Personalized element:} .\\
\textbf{Personalized info:} name.
\label{box2}

\vspace{4pt}
\hrulefill
\vspace{4pt}

\textbf{Instruction ID:} 3.\\
\textbf{Natural-language instruction:} Create a birthday schedule for May 20th.\\
\textbf{Difficulty label:} easy.\\
\textbf{The minimal step count:} 6.\\
\textbf{The app(s) involved:} Calendar.\\
\textbf{The completed form of instruction:} Create a birthday schedule for \{date\}.\\
\textbf{Personalized element:} .\\
\textbf{Personalized info:} date.
\label{box3}

\vspace{4pt}
\hrulefill
\vspace{4pt}

\textbf{Instruction ID:} 4.\\
\textbf{Natural-language instruction:} Open the settings to unload less commonly used network disks app.\\
\textbf{Difficulty label:} easy.\\
\textbf{The minimal step count:} 7.\\
\textbf{The app(s) involved:} Settings.\\
\textbf{The completed form of instruction:} Open the settings to unload \{app name\}.\\
\textbf{Personalized element:} less commonly used network disks app.\\
\textbf{Personalized info:} app name.
\label{box4}

\vspace{4pt}
\hrulefill
\vspace{4pt}

\textbf{Instruction ID:} 5.\\
\textbf{Natural-language instruction:} Set an alarm for 7:30 in the morning.\\
\textbf{Difficulty label:} easy.\\
\textbf{The minimal step count:} 5.\\
\textbf{The app(s) involved:} Clock.\\
\textbf{The completed form of instruction:} Set an alarm for \{time\} in the morning.\\
\textbf{Personalized element:} .\\
\textbf{Personalized info:} time.
\label{box5}

\vspace{4pt}
\hrulefill
\vspace{4pt}

\textbf{Instruction ID:} 6.\\
\textbf{Natural-language instruction:} Open the bilibili to praise the latest video of the favorite up.\\
\textbf{Difficulty label:} easy.\\
\textbf{The minimal step count:} 6.\\
\textbf{The app(s) involved:} Bilibili.\\
\textbf{The completed form of instruction:} Open the bilibili to praise the latest video of \{up name\}.\\
\textbf{Personalized element:} Favorite up.\\
\textbf{Personalized info:} up name.
\label{box6}

\vspace{4pt}
\hrulefill
\vspace{4pt}

\textbf{Instruction ID:} 7.\\
\textbf{Natural-language instruction:} Play the song ‘Something Just Like This'.\\
\textbf{Difficulty label:} easy.\\
\textbf{The minimal step count:} 5.\\
\textbf{The app(s) involved:} NetEase Cloud Music.\\
\textbf{The completed form of instruction:} Play the song \{song name\}.\\
\textbf{Personalized element:} .\\
\textbf{Personalized info:} song name.
\label{box7}

\vspace{4pt}
\hrulefill
\vspace{4pt}

\textbf{Instruction ID:} 8.\\
\textbf{Natural-language instruction:} Take a taxi to New York.\\
\textbf{Difficulty label:} easy.\\
\textbf{The minimal step count:} 6.\\
\textbf{The app(s) involved:} Didi Chuxing.\\
\textbf{The completed form of instruction:} Take a taxi to \{place name\}.\\
\textbf{Personalized element:} .\\
\textbf{Personalized info:} place name.
\label{box8}

\vspace{4pt}
\hrulefill
\vspace{4pt}

\textbf{Instruction ID:} 9.\\
\textbf{Natural-language instruction:} Open Pinduoduo to collect one often bought snack.\\
\textbf{Difficulty label:} easy.\\
\textbf{The minimal step count:} 6.\\
\textbf{The app(s) involved:} Pinduoduo.\\
\textbf{The completed form of instruction:} Open Pinduoduo to collect one \{snack name\}.\\
\textbf{Personalized element:} often bought snack.\\
\textbf{Personalized info:} snack name.
\label{box9}

\vspace{4pt}
\hrulefill
\vspace{4pt}

\textbf{Instruction ID:} 10.\\
\textbf{Natural-language instruction:} Order a dumping takeout.\\
\textbf{Difficulty label:} easy.\\
\textbf{The minimal step count:} 7.\\
\textbf{The app(s) involved:} elm.me.me.\\
\textbf{The completed form of instruction:} Order a \{food name\} takeout.\\
\textbf{Personalized element:} .\\
\textbf{Personalized info:} food name.
\label{box10}

\vspace{4pt}
\hrulefill
\vspace{4pt}

\textbf{Instruction ID:} 11.\\
\textbf{Natural-language instruction:} Open WeChat to reply hello to friend.\\
\textbf{Difficulty label:} easy.\\
\textbf{The minimal step count:} 5.\\
\textbf{The app(s) involved:} WeChat.\\
\textbf{The completed form of instruction:} Open WeChat to reply hello to \{name\}.\\
\textbf{Personalized element:} Friend.\\
\textbf{Personalized info:} name.
\label{box11}

\vspace{4pt}
\hrulefill
\vspace{4pt}

\textbf{Instruction ID:} 12.\\
\textbf{Natural-language instruction:} Search for fast food restaurants near my home through browser and click on the first one.\\
\textbf{Difficulty label:} easy.\\
\textbf{The minimal step count:} 5.\\
\textbf{The app(s) involved:} Browser.\\
\textbf{The completed form of instruction:} Search for fast food restaurants near \{place name\} through browser and click on the first one.\\
\textbf{Personalized element:}my home.\\
\textbf{Personalized info:} place name.
\label{box12}

\vspace{4pt}
\hrulefill
\vspace{4pt}

\textbf{Instruction ID:} 13.\\
\textbf{Natural-language instruction:} Navigate to New York with walking.\\
\textbf{Difficulty label:} easy.\\
\textbf{The minimal step count:} 6.\\
\textbf{The app(s) involved:} Baidumap.\\
\textbf{The completed form of instruction:} Navigate to \{place name\} with walking.\\
\textbf{Personalized element:} .\\
\textbf{Personalized info:} place name.
\label{box13}

\vspace{4pt}
\hrulefill
\vspace{4pt}

\textbf{Instruction ID:} 14.\\
\textbf{Natural-language instruction:} Send a text message to jack that your have worked hard.\\
\textbf{Difficulty label:} easy.\\
\textbf{The minimal step count:} 8.\\
\textbf{The app(s) involved:} Messages.\\
\textbf{The completed form of instruction:} Send a text message to \{name\} that your have worked hard.\\
\textbf{Personalized element:} .\\
\textbf{Personalized info:} name.
\label{box14}

\vspace{4pt}
\hrulefill
\vspace{4pt}

\textbf{Instruction ID:} 15.\\
\textbf{Natural-language instruction:} Open rednote check a post on the list of the often bought snack and point out one love.\\
\textbf{Difficulty label:} easy.\\
\textbf{The minimal step count:} 7.\\
\textbf{The app(s) involved:} rednote.\\
\textbf{The completed form of instruction:} Open rednote check a post on the list of \{snack name\} and point out one love.\\
\textbf{Personalized element:} often bought snack.\\
\textbf{Personalized info:} snack name.
\label{box15}

\vspace{4pt}
\hrulefill
\vspace{4pt}

\textbf{Instruction ID:} 16.\\
\textbf{Natural-language instruction:} Open Jingdong to buy own computer again.\\
\textbf{Difficulty label:} easy.\\
\textbf{The minimal step count:} 8.\\
\textbf{The app(s) involved:} Jingdong.\\
\textbf{The completed form of instruction:} Open Jingdong to buy \{computer name\} again.\\
\textbf{Personalized element:} Own computer.\\
\textbf{Personalized info:} computer name.
\label{box16}

\vspace{4pt}
\hrulefill
\vspace{4pt}

\textbf{Instruction ID:} 17.\\
\textbf{Natural-language instruction:} Open deepseek for 100-word information about my school.\\
\textbf{Difficulty label:} easy.\\
\textbf{The minimal step count:} 4.\\
\textbf{The app(s) involved:} DeepSeek.\\
\textbf{The completed form of instruction:} Open deepseek for 100-word information about \{school name\}.\\
\textbf{Personalized element:} my school.\\
\textbf{Personalized info:} school name.
\label{box17}

\vspace{4pt}
\hrulefill
\vspace{4pt}

\textbf{Instruction ID:} 18.\\
\textbf{Natural-language instruction:} Open WeChat to scan the QR code sent by friend.\\
\textbf{Difficulty label:} easy.\\
\textbf{The minimal step count:} 5.\\
\textbf{The app(s) involved:} WeChat.\\
\textbf{The completed form of instruction:} Open WeChat to scan the QR code sent by \{name\}.\\
\textbf{Personalized element:} Friend.\\
\textbf{Personalized info:} name.
\label{box18}

\vspace{4pt}
\hrulefill
\vspace{4pt}

\textbf{Instruction ID:} 19.\\
\textbf{Natural-language instruction:} Transfer 100 yuan to Bob.\\
\textbf{Difficulty label:} easy.\\
\textbf{The minimal step count:} 8.\\
\textbf{The app(s) involved:} WeChat.\\
\textbf{The completed form of instruction:} Transfer 100 yuan to \{name\}.\\
\textbf{Personalized element:} .\\
\textbf{Personalized info:} name.
\label{box19}

\vspace{4pt}
\hrulefill
\vspace{4pt}

\textbf{Instruction ID:} 20.\\
\textbf{Natural-language instruction:} Open QQ to check the questions sent by friend and copy it.\\
\textbf{Difficulty label:} easy.\\
\textbf{The minimal step count:} 4.\\
\textbf{The app(s) involved:} QQ.\\
\textbf{The completed form of instruction:} Open QQ to check the questions sent by \{name\} and copy it.\\
\textbf{Personalized element:} friend.\\
\textbf{Personalized info:} name.
\label{box20}

\vspace{4pt}
\hrulefill
\vspace{4pt}

\textbf{Instruction ID:} 21.\\
\textbf{Natural-language instruction:} Open the railway 12306 to check the high-speed train ticket from Chongqing to my city at 6:42 tomorrow.\\
\textbf{Difficulty label:} easy.\\
\textbf{The minimal step count:} 8.\\
\textbf{The app(s) involved:} railway 12306.\\
\textbf{The completed form of instruction:} Open the railway 12306 to check the high-speed train ticket from Chongqing to \{city name\} at 6:42 tomorrow.\\
\textbf{Personalized element:} my city.\\
\textbf{Personalized info:} city name.
\label{box21}

\vspace{4pt}
\hrulefill
\vspace{4pt}

\textbf{Instruction ID:} 22.\\
\textbf{Natural-language instruction:} Open QQ to reply hello to friend and add a often used expression.\\
\textbf{Difficulty label:} normal.\\
\textbf{The minimal step count:} 7.\\
\textbf{The app(s) involved:} QQ.\\
\textbf{The completed form of instruction:} Open QQ to reply hello to \{name\} and add a \{emoji\}.\\
\textbf{Personalized element:} friend, often used expression.\\
\textbf{Personalized info:} name, emoji.
\label{box22}

\vspace{4pt}
\hrulefill
\vspace{4pt}

\textbf{Instruction ID:} 23.\\
\textbf{Natural-language instruction:} Open NetEase Cloud Music to play favorite song and forward it to friend.\\
\textbf{Difficulty label:} normal.\\
\textbf{The minimal step count:} 9.\\
\textbf{The app(s) involved:} NetEase Cloud Music.\\
\textbf{The completed form of instruction:} Open NetEase Cloud Music to play \{song name\} and forward it to \{name\}.\\
\textbf{Personalized element:} Favorite song, friend.\\
\textbf{Personalized info:} song name, name.
\label{box23}

\vspace{4pt}
\hrulefill
\vspace{4pt}

\textbf{Instruction ID:} 24.\\
\textbf{Natural-language instruction:} Open the elm.me and order a frequent takeout and the address to fill in my brother's home.\\
\textbf{Difficulty label:} normal.\\
\textbf{The minimal step count:} 11.\\
\textbf{The app(s) involved:} elm.me.\\
\textbf{The completed form of instruction:} Open the elm.me and order a \{food name\} and the address to fill in \{address\}.\\
\textbf{Personalized element:} Frequent takeout, my Brother's home.\\
\textbf{Personalized info:} food name, address.
\label{box24}

\vspace{4pt}
\hrulefill
\vspace{4pt}

\textbf{Instruction ID:} 25.\\
\textbf{Natural-language instruction:} Open the railway 12306 and buy friend a high-speed train ticket from Chongqing to my city tomorrow.\\
\textbf{Difficulty label:} normal.\\
\textbf{The minimal step count:} 11.\\
\textbf{The app(s) involved:} railway 12306.\\
\textbf{The completed form of instruction:} Open the railway 12306 and buy \{name\} a high-speed train ticket from Chongqing to the \{city name\} tomorrow.\\
\textbf{Personalized element:} my city, School.\\
\textbf{Personalized info:} name, city name.
\label{box25}

\vspace{4pt}
\hrulefill
\vspace{4pt}

\textbf{Instruction ID:} 26.\\
\textbf{Natural-language instruction:} Open Weibo to check favorite star and forward his latest developments to the friend.\\
\textbf{Difficulty label:} normal.\\
\textbf{The minimal step count:} 8.\\
\textbf{The app(s) involved:} Weibo.\\
\textbf{The completed form of instruction:} Open Weibo to check \{star name\} and forward his latest developments to \{name\}.\\
\textbf{Personalized element:} Favorite star, Friend.\\
\textbf{Personalized info:} star name, name.
\label{box26}

\vspace{4pt}
\hrulefill
\vspace{4pt}

\textbf{Instruction ID:} 27.\\
\textbf{Natural-language instruction:} Open the cainiao app and add a new harvest address, fill in the friend, fill in the Friend's phone number, and the address is the first one to locate.\\
\textbf{Difficulty label:} normal.\\
\textbf{The minimal step count:} 9.\\
\textbf{The app(s) involved:} Cainiao.\\
\textbf{The completed form of instruction:} Open the cainiao app and add a new harvest address, fill in \{name\}, fill in \{phone number\}, and the address is the first one to locate.\\
\textbf{Personalized element:} Friend's phone number, Friend.\\
\textbf{Personalized info:} name, phone number.
\label{box27}

\vspace{4pt}
\hrulefill
\vspace{4pt}

\textbf{Instruction ID:} 28.\\
\textbf{Natural-language instruction:} Open QQ to forward professional group's latest information to friend.\\
\textbf{Difficulty label:} normal.\\
\textbf{The minimal step count:} 6.\\
\textbf{The app(s) involved:} QQ.\\
\textbf{The completed form of instruction:} Open QQ to forward \{group name\}'s latest information to \{name\}.\\
\textbf{Personalized element:} Professional group, friend.\\
\textbf{Personalized info:} group name, name.
\label{box28}

\vspace{4pt}
\hrulefill
\vspace{4pt}

\textbf{Instruction ID:} 29.\\
\textbf{Natural-language instruction:} Open Jingdong to buy the own computer again. The consignee is friend.\\
\textbf{Difficulty label:} normal.\\
\textbf{The minimal step count:} 10.\\
\textbf{The app(s) involved:} Jingdong.\\
\textbf{The completed form of instruction:} Open Jingdong to buy \{computer name\} again. The consignee is \{name\}.\\
\textbf{Personalized element:} own computer, friend.\\
\textbf{Personalized info:} computer name, name.
\label{box29}

\vspace{4pt}
\hrulefill
\vspace{4pt}

\textbf{Instruction ID:} 30.\\
\textbf{Natural-language instruction:} open TikTok to search for my school and then forward the first video to TikTok friend with comment good-looking videos.\\
\textbf{Difficulty label:} normal.\\
\textbf{The minimal step count:} 11.\\
\textbf{The app(s) involved:} TikTok.\\
\textbf{The completed form of instruction:} open TikTok to search for \{school name\} and then forward the first video to \{name\} with comment good-looking videos.\\
\textbf{Personalized element:} my School, TikTok friend.\\
\textbf{Personalized info:} school name, name.
\label{box30}

\vspace{4pt}
\hrulefill
\vspace{4pt}

\textbf{Instruction ID:} 31.\\
\textbf{Natural-language instruction:} Open Taobao to buy a Collectible doll for girlfriend and then note Happy Valentine's Day.\\
\textbf{Difficulty label:} normal.\\
\textbf{The minimal step count:} 14.\\
\textbf{The app(s) involved:} Taobao.\\
\textbf{The completed form of instruction:} Open Taobao to buy a \{doll name\} for \{name\} and then note Happy Valentine's Day.\\
\textbf{Personalized element:} girlfriend, Collectible doll.\\
\textbf{Personalized info:} doll name, name.
\label{box31}

\vspace{4pt}
\hrulefill
\vspace{4pt}

\textbf{Instruction ID:} 32.\\
\textbf{Natural-language instruction:} Set an alarm for The start time of the class and then set an alarm for the end time of the class.\\
\textbf{Difficulty label:} normal.\\
\textbf{The minimal step count:} 10.\\
\textbf{The app(s) involved:} Clock.\\
\textbf{The completed form of instruction:} Set an alarm for \{time\} and then set an alarm for \{time\}.\\
\textbf{Personalized element:} start time of the class, end time of the class.\\
\textbf{Personalized info:} time, time.
\label{box32}

\vspace{4pt}
\hrulefill
\vspace{4pt}

\textbf{Instruction ID:} 33.\\
\textbf{Natural-language instruction:} Open the text message and send a message to mom that you have worked hard, then check the text message sent by friend and reply to the receipt.\\
\textbf{Difficulty label:} normal.\\
\textbf{The minimal step count:} 13.\\
\textbf{The app(s) involved:} Messages.\\
\textbf{The completed form of instruction:} Open the text message and send a message to \{name\} that you have worked hard, then check the text message sent by \{name\} and reply to the receipt.\\
\textbf{Personalized element:} Mom, Friend.\\
\textbf{Personalized info:} name, name.
\label{box33}

\vspace{4pt}
\hrulefill
\vspace{4pt}

\textbf{Instruction ID:} 34.\\
\textbf{Natural-language instruction:} Open WeChat to scan the QRcode sent by friend and send the information to Own WeChat.\\
\textbf{Difficulty label:} normal.\\
\textbf{The minimal step count:} 12.\\
\textbf{The app(s) involved:} WeChat.\\
\textbf{The completed form of instruction:} Open WeChat to scan the QRcode sent by \{name\} and send the information to \{wechat name\}.\\
\textbf{Personalized element:} Friend, own wechat.\\
\textbf{Personalized info:} name, wechat name.
\label{box34}

\vspace{4pt}
\hrulefill
\vspace{4pt}

\textbf{Instruction ID:} 35.\\
\textbf{Natural-language instruction:} Open Baidu map, navigate to my home and then open QQ to send a message to friend that I am almost Home.\\
\textbf{Difficulty label:} difficult.\\
\textbf{The minimal step count:} 12.\\
\textbf{The app(s) involved:} Baidu Map, QQ.\\
\textbf{The completed form of instruction:} Open Baidu map, navigate to \{place name\} and then open QQ to send a message to \{name\} that I am almost Home.\\
\textbf{Personalized element:} Home, Friend.\\
\textbf{Personalized info:} place name, name.
\label{box35}

\vspace{4pt}
\hrulefill
\vspace{4pt}

\textbf{Instruction ID:} 36.\\
\textbf{Natural-language instruction:} Turn on Didi Chuxing to take a taxi to my home, and then open QQ to tell friend that I am coming by taxi.\\
\textbf{Difficulty label:} difficult.\\
\textbf{The minimal step count:} 12.\\
\textbf{The app(s) involved:} Didi Chuxing, QQ.\\
\textbf{The completed form of instruction:} Turn on Didi Chuxing to take a taxi to \{place name\}, and then open QQ to tell \{name\} that I am coming by taxi.\\
\textbf{Personalized element:} my home, friend.\\
\textbf{Personalized info:} place name, name.
\label{box36}

\vspace{4pt}
\hrulefill
\vspace{4pt}

\textbf{Instruction ID:} 37.\\
\textbf{Natural-language instruction:} open Rednote check one list of often bought snack and collect it and then forward it to friend via WeChat.\\
\textbf{Difficulty label:} difficult.\\
\textbf{The minimal step count:} 11.\\
\textbf{The app(s) involved:} rednote, WeChat.\\
\textbf{The completed form of instruction:} open Rednote check one list of \{snack name\} and collect it and then forward it to \{name\} via WeChat.\\
\textbf{Personalized element:} Friend, often bought snack.\\
\textbf{Personalized info:} snack name, name.
\label{box37}

\vspace{4pt}
\hrulefill
\vspace{4pt}

\textbf{Instruction ID:} 38.\\
\textbf{Natural-language instruction:} Open deepseek to ask the my school's 100-word message and send it to friend via QQ.\\
\textbf{Difficulty label:} difficult.\\
\textbf{The minimal step count:} 11.\\
\textbf{The app(s) involved:} DeepSeek, QQ.\\
\textbf{The completed form of instruction:} Open deepseek to ask \{school name\}'s 100-word message and send it to \{name\} via QQ.\\
\textbf{Personalized element:} my School, friend.\\
\textbf{Personalized info:} school name, name.
\label{box38}

\vspace{4pt}
\hrulefill
\vspace{4pt}

\textbf{Instruction ID:} 39.\\
\textbf{Natural-language instruction:} open Jingdong to Check the selling price of own computer and send it to friend via QQ.\\
\textbf{Difficulty label:} difficult.\\
\textbf{The minimal step count:} 11.\\
\textbf{The app(s) involved:} Jingdong, QQ.\\
\textbf{The completed form of instruction:} open Jingdong to Check the selling price of \{computer name\} and send it to \{name\} via QQ.\\
\textbf{Personalized element:} own computer, Friend.\\
\textbf{Personalized info:} computer name, name.
\label{box39}

\vspace{4pt}
\hrulefill
\vspace{4pt}

\textbf{Instruction ID:} 40.\\
\textbf{Natural-language instruction:} open the settings to connect to the dormitory's WiFi and then turn on the tiktok to search for my school and tap one video.\\
\textbf{Difficulty label:} difficult.\\
\textbf{The minimal step count:} 9.\\
\textbf{The app(s) involved:} Settings, TikTok.\\
\textbf{The completed form of instruction:} open the settings to connect to \{wifi name\} and then turn on the tiktok to search for \{school name\} and tap one video.\\
\textbf{Personalized element:} Dormitory WiFi, my School.\\
\textbf{Personalized info:} wifi name, school name.
\label{box40}

\vspace{4pt}
\hrulefill
\vspace{4pt}

\textbf{Instruction ID:} 41.\\
\textbf{Natural-language instruction:} open the clock and set the alarm clock with the start time of the class and the end time of the class, then send the class time to friend via QQ.\\
\textbf{Difficulty label:} difficult.\\
\textbf{The minimal step count:} 10.\\
\textbf{The app(s) involved:} Clock, QQ.\\
\textbf{The completed form of instruction:} open the clock and set the alarm clock with \{time\} and \{time\}, then send the class time to \{name\} via QQ.\\
\textbf{Personalized element:} start time of the class, end time of the class, Friend.\\
\textbf{Personalized info:} time, time, name.
\label{box41}

\vspace{4pt}
\hrulefill
\vspace{4pt}

\textbf{Instruction ID:} 42.\\
\textbf{Natural-language instruction:} open rednote to check a post on the list of the often bought snack and point out one love, then forward it to Friend on WeChat and add it to collection on Pinduoduo.\\
\textbf{Difficulty label:} difficult.\\
\textbf{The minimal step count:} 16.\\
\textbf{The app(s) involved:} rednote, WeChat, Pinduoduo.\\
\textbf{The completed form of instruction:} open rednote to check a post on the list of \{snack name\} and point out one love, then forward it to \{name\} on WeChat and add it to collection on Pinduoduo.\\
\textbf{Personalized element:} often bought snack, Friend.\\
\textbf{Personalized info:} snack name, name.
\label{box42}

\vspace{4pt}
\hrulefill
\vspace{4pt}

\textbf{Instruction ID:} 43.\\
\textbf{Natural-language instruction:} open bilibili to Check the video of favorite upmaster and praise it and then forward it to friend via WeChat.\\
\textbf{Difficulty label:} difficult.\\
\textbf{The minimal step count:} 10.\\
\textbf{The app(s) involved:} Bilibili, WeChat.\\
\textbf{The completed form of instruction:} open bilibili to Check the video of \{up name\} and praise it and then forward it to \{name\} via WeChat.\\
\textbf{Personalized element:} Favorite upmaster, Friend.\\
\textbf{Personalized info:} up name, name.
\label{box43}

\vspace{4pt}
\hrulefill
\vspace{4pt}

\textbf{Instruction ID:} 44.\\
\textbf{Natural-language instruction:} Set an alarm clock at the end time of the class and use the elm.me to order a Frequent takeout.\\
\textbf{Difficulty label:} difficult.\\
\textbf{The minimal step count:} 14.\\
\textbf{The app(s) involved:} Clock, elm.me.\\
\textbf{The completed form of instruction:} Set an alarm clock at \{time\} and use the elm.me to order a \{food name\}.\\
\textbf{Personalized element:} the end of time of the class, Frequent takeout.\\
\textbf{Personalized info:} time, food name.
\label{box44}

\vspace{4pt}
\hrulefill
\vspace{4pt}

\textbf{Instruction ID:} 45.\\
\textbf{Natural-language instruction:} Open the railway 12306 to check the high-speed train ticket from Chongqing to my city at 6:42 tomorrow and then open QQ to send the arrival time to friend.\\
\textbf{Difficulty label:} difficult.\\
\textbf{The minimal step count:} 14.\\
\textbf{The app(s) involved:} railway 12306, QQ.\\
\textbf{The completed form of instruction:} Open the railway 12306 to check the high-speed train ticket from Chongqing to \{city name\} at 6:42 tomorrow and then open QQ to send the arrival time to \{name\}.\\
\textbf{Personalized element:} my city, Friend.\\
\textbf{Personalized info:} city name, name.
\label{box45}

\vspace{4pt}
\hrulefill
\vspace{4pt}

\textbf{Instruction ID:} 46.\\
\textbf{Natural-language instruction:} Open the calendar and set a birthday schedule on the day of Dad's birthday, then open the phone to call Mom and then open QQ to send a message to brother that the content is the date of Dad's birthday.\\
\textbf{Difficulty label:} difficult.\\
\textbf{The minimal step count:} 17.\\
\textbf{The app(s) involved:} Calendar, Phone, QQ.\\
\textbf{The completed form of instruction:} Open the calendar and set a birthday schedule on \{date\}, then open the phone to call \{name\} and then open QQ to send a message to \{name\} that the content is the date of Dad's birthday.\\
\textbf{Personalized element:} Dad's birthday, Mother, Brother.\\
\textbf{Personalized info:} date, name, name.
\label{box46}

\vspace{4pt}
\hrulefill
\vspace{4pt}

\textbf{Instruction ID:} 47.\\
\textbf{Natural-language instruction:} Open Jingdong to buy own computer again. The consignee is friend and then Open WeChat and send a message to friend that have bought a computer.\\
\textbf{Difficulty label:} difficult.\\
\textbf{The minimal step count:} 16.\\
\textbf{The app(s) involved:} Jingdong, WeChat.\\
\textbf{The completed form of instruction:} Open Jingdong to buy \{computer name\} again. The consignee is \{name\} and then Open WeChat and send a message to \{name\} that have bought a computer.\\
\textbf{Personalized element:} own computer, friend.\\
\textbf{Personalized info:} computer name, name, name.
\label{box47}

\vspace{4pt}
\hrulefill
\vspace{4pt}

\textbf{Instruction ID:} 48.\\
\textbf{Natural-language instruction:} Open NetEase Cloud Music Search and play the favorite song and then forward it to friend and then open browser to search for the creation background of the song.\\
\textbf{Difficulty label:} difficult.\\
\textbf{The minimal step count:} 15.\\
\textbf{The app(s) involved:} NetEase Cloud Music, Browser.\\
\textbf{The completed form of instruction:} Open NetEase Cloud Music Search and play \{song name\} and then forward it to \{name\} and then open browser to search for the creation background of the song.\\
\textbf{Personalized element:} favorite song, Friend.\\
\textbf{Personalized info:} song name, name.
\label{box48}

\vspace{4pt}
\hrulefill
\vspace{4pt}

\textbf{Instruction ID:} 49.\\
\textbf{Natural-language instruction:} Open the text message and send a message to mother that you have worked hard, then check the text message sent by friend and reply to receive it, then open QQ and send the content of friend's text message to brother.\\
\textbf{Difficulty label:} difficult.\\
\textbf{The minimal step count:} 19.\\
\textbf{The app(s) involved:} Messages, QQ.\\
\textbf{The completed form of instruction:} Open the text message and send a message to \{name\} that you have worked hard, then check the text message sent by \{name\} and reply to receive it, then open QQ and send the content of friend's text message to \{name\}.\\
\textbf{Personalized element:} Mom, Brother, Friend.\\
\textbf{Personalized info:} name, name, name.
\label{box49}

\vspace{4pt}
\hrulefill
\vspace{4pt}

\textbf{Instruction ID:} 50.\\
\textbf{Natural-language instruction:} Open QQ to view the latest news of friend, then follow the settings of this message to open the operation, and then send the completed operation in the professional group.\\
\textbf{Difficulty label:} difficult.\\
\textbf{The minimal step count:} 21.\\
\textbf{The app(s) involved:} QQ.\\
\textbf{The completed form of instruction:} Open QQ to view the latest news of \{name\}, then follow the settings of this message to open the operation, and then send the completed operation in the \{group name\}.\\
\textbf{Personalized element:} Friend, Professional Group.\\
\textbf{Personalized info:} name, group name.
\label{box50}

\vspace{4pt}
\hrulefill
\vspace{4pt}

\textbf{Instruction ID:} 51.\\
\textbf{Natural-language instruction:} Open QQ to check your friend's taboo, then open elm.me order the Frequent takeout, and then comment on friend's taboo.\\
\textbf{Difficulty label:} difficult.\\
\textbf{The minimal step count:} 15.\\
\textbf{The app(s) involved:} QQ, elm.me.\\
\textbf{The completed form of instruction:} Open QQ to check \{taboo\}, then open elm.me order \{food name\}, and then comment on friend's taboo.\\
\textbf{Personalized element:} Friend's taboo, Frequent takeout.\\
\textbf{Personalized info:} taboo, food name.
\label{box51}

\vspace{4pt}
\hrulefill
\vspace{4pt}

\textbf{Instruction ID:} 52.\\
\textbf{Natural-language instruction:} Open WeChat to scan the QRcode sent by friend and then send the information to Own WeChat. Open QQ to send the information to professional group.\\
\textbf{Difficulty label:} difficult.\\
\textbf{The minimal step count:} 19.\\
\textbf{The app(s) involved:} WeChat, QQ.\\
\textbf{The completed form of instruction:} Open WeChat to scan the QRcode sent by \{name\} and then send the information to \{wechat name\}. Open QQ to send the information to \{group name\}.\\
\textbf{Personalized element:} Friend, Own WeChat, Professional group.\\
\textbf{Personalized info:} name, wechat name, group name.
\label{box52}

\vspace{4pt}
\hrulefill
\vspace{4pt}

\textbf{Instruction ID:} 53.\\
\textbf{Natural-language instruction:} Open Baidu map, navigate to my home, then open QQ to send a message to friend that I am almost home, and then send a message in QQ space that I am Home.\\
\textbf{Difficulty label:} difficult.\\
\textbf{The minimal step count:} 19.\\
\textbf{The app(s) involved:} Baidu Map, QQ.\\
\textbf{The completed form of instruction:} Open Baidu map, navigate to \{place name\}, then open QQ to send a message to \{name\} that I am almost home, and then send a message in QQ space that I am Home.\\
\textbf{Personalized element:} Home, Friend.\\
\textbf{Personalized info:} place name, name.
\label{box53}

\vspace{4pt}
\hrulefill
\vspace{4pt}

\textbf{Instruction ID:} 54.\\
\textbf{Natural-language instruction:} Open QQ to check the questions sent by friend, then open deepseek to ask this question and then open QQ send the answer to professional group.\\
\textbf{Difficulty label:} difficult.\\
\textbf{The minimal step count:} 18.\\
\textbf{The app(s) involved:} QQ, DeepSeek.\\
\textbf{The completed form of instruction:} Open QQ to check the questions sent by \{name\}, then open deepseek to ask this question and then open QQ send the answer to \{group name\}.\\
\textbf{Personalized element:} Friend, Professional Groups.\\
\textbf{Personalized info:} name, group name.
\label{box54}

\vspace{4pt}
\hrulefill
\vspace{4pt}

\textbf{Instruction ID:} 55.\\
\textbf{Natural-language instruction:} Open the railway 12306 to check the high-speed train tickets from Chongqing 6.42 to my city tomorrow, then open QQ to send friend the arrival time of the high-speed train and then open the browser to inquire about the scenic spots near my city.\\
\textbf{Difficulty label:} difficult.\\
\textbf{The minimal step count:} 20.\\
\textbf{The app(s) involved:} railway 12306, QQ, Browser.\\
\textbf{The completed form of instruction:} Open the railway 12306 to check the high-speed train tickets from Chongqing 6.42 to \{city name\} tomorrow, then open QQ to send \{name\} the arrival time of the high-speed train and then open the browser to inquire about the scenic spots near \{city name\}.\\
\textbf{Personalized element:} my city, Friend.\\
\textbf{Personalized info:} city name, name, city name.
\label{box55}

\vspace{4pt}
\hrulefill
\vspace{4pt}

\textbf{Instruction ID:} 56.\\
\textbf{Natural-language instruction:} Open the TikTok to publish a text video with the text hello world and then @ TikTok friend.\\
\textbf{Difficulty label:} easy.\\
\textbf{The minimal step count:} 8.\\
\textbf{The app(s) involved:} TikTok.\\
\textbf{The completed form of instruction:} Open the TikTok to publish a text video with the text hello world and then @\{name\}.\\
\textbf{Personalized element:} TikTok Friend.\\
\textbf{Personalized info:} name.
\label{box56}

\vspace{4pt}
\hrulefill
\vspace{4pt}

\textbf{Instruction ID:} 57.\\
\textbf{Natural-language instruction:} Open the bilibili to play the collected animation video and praise it.\\
\textbf{Difficulty label:} easy.\\
\textbf{The minimal step count:} 5.\\
\textbf{The app(s) involved:} Bilibili.\\
\textbf{The completed form of instruction:} Open the bilibili to play \{video name\} and praise it.\\
\textbf{Personalized element:} Collection of anime videos.\\
\textbf{Personalized info:} video name.
\label{box57}

\vspace{4pt}
\hrulefill
\vspace{4pt}

\textbf{Instruction ID:} 58.\\
\textbf{Natural-language instruction:} Open the weather to check the weather of my school tomorrow.\\
\textbf{Difficulty label:} easy.\\
\textbf{The minimal step count:} 7.\\
\textbf{The app(s) involved:} Weather.\\
\textbf{The completed form of instruction:} Open the weather to check the weather of \{school name\} tomorrow.\\
\textbf{Personalized element:} my school.\\
\textbf{Personalized info:} school name.
\label{box58}

\vspace{4pt}
\hrulefill
\vspace{4pt}

\textbf{Instruction ID:} 59.\\
\textbf{Natural-language instruction:} Open rednote, search for articles in the research direction,  click on the first one and click on the heart.\\
\textbf{Difficulty label:} easy.\\
\textbf{The minimal step count:} 7.\\
\textbf{The app(s) involved:} rednote.\\
\textbf{The completed form of instruction:} Open rednote, search for articles in \{research direction\},  click on the first one and click on the heart.\\
\textbf{Personalized element:} research direction.\\
\textbf{Personalized info:} research direction.
\label{box59}

\vspace{4pt}
\hrulefill
\vspace{4pt}

\textbf{Instruction ID:} 60.\\
\textbf{Natural-language instruction:} Open Taobao search for school's merchandise.\\
\textbf{Difficulty label:} easy.\\
\textbf{The minimal step count:} 4.\\
\textbf{The app(s) involved:} Taobao.\\
\textbf{The completed form of instruction:} Open Taobao search for \{school name\}'s merchandise.\\
\textbf{Personalized element:} school.\\
\textbf{Personalized info:} school name.
\label{box60}

\vspace{4pt}
\hrulefill
\vspace{4pt}

\textbf{Instruction ID:} 61.\\
\textbf{Natural-language instruction:} Open Browser search for school's pictures and save one.\\
\textbf{Difficulty label:} easy.\\
\textbf{The minimal step count:} 7.\\
\textbf{The app(s) involved:} Browser.\\
\textbf{The completed form of instruction:} Open Browser search for \{school name\}'s pictures and save one.\\
\textbf{Personalized element:} school.\\
\textbf{Personalized info:} school name.
\label{box61}

\vspace{4pt}
\hrulefill
\vspace{4pt}

\textbf{Instruction ID:} 62.\\
\textbf{Natural-language instruction:} Open Settings to connect Dormitory WiFi.\\
\textbf{Difficulty label:} easy.\\
\textbf{The minimal step count:} 4.\\
\textbf{The app(s) involved:} Settings.\\
\textbf{The completed form of instruction:} Open Settings to connect \{wifi name\}.\\
\textbf{Personalized element:} Dormitory WiFi.\\
\textbf{Personalized info:} wifi name.
\label{box62}

\vspace{4pt}
\hrulefill
\vspace{4pt}

\textbf{Instruction ID:} 63.\\
\textbf{Natural-language instruction:} Open QQ to download videos from professional group.\\
\textbf{Difficulty label:} easy.\\
\textbf{The minimal step count:} 4.\\
\textbf{The app(s) involved:} QQ.\\
\textbf{The completed form of instruction:} Open QQ to download videos from \{group name\}.\\
\textbf{Personalized element:} professional group.\\
\textbf{Personalized info:} group name.
\label{box63}

\vspace{4pt}
\hrulefill
\vspace{4pt}

\textbf{Instruction ID:} 64.\\
\textbf{Natural-language instruction:} Open WPS to create a blank document, enter friend's phone number and save it to the cloud disk.\\
\textbf{Difficulty label:} easy.\\
\textbf{The minimal step count:} 8.\\
\textbf{The app(s) involved:} WPS.\\
\textbf{The completed form of instruction:} Open WPS to create a blank document, enter \{phone number\} and save it to the cloud disk.\\
\textbf{Personalized element:} friend's phone number.\\
\textbf{Personalized info:} phone number.
\label{box64}

\vspace{4pt}
\hrulefill
\vspace{4pt}

\textbf{Instruction ID:} 65.\\
\textbf{Natural-language instruction:} Open App Market to download less commonly used network disks app.\\
\textbf{Difficulty label:} easy.\\
\textbf{The minimal step count:} 5.\\
\textbf{The app(s) involved:} App Market.\\
\textbf{The completed form of instruction:} Open App Market to download \{app name\}.\\
\textbf{Personalized element:} less commonly used network disks app.\\
\textbf{Personalized info:} app name.
\label{box65}

\vspace{4pt}
\hrulefill
\vspace{4pt}

\textbf{Instruction ID:} 66.\\
\textbf{Natural-language instruction:} Open the Dianping Search for tourist routes near the school and click on the first.\\
\textbf{Difficulty label:} easy.\\
\textbf{The minimal step count:} 6.\\
\textbf{The app(s) involved:} Dianping.\\
\textbf{The completed form of instruction:} Open the Dianping Search for tourist routes near \{school name\} and click on the first.\\
\textbf{Personalized element:} school.\\
\textbf{Personalized info:} school name.
\label{box66}

\vspace{4pt}
\hrulefill
\vspace{4pt}

\textbf{Instruction ID:} 67.\\
\textbf{Natural-language instruction:} Open the TikTok to publish a text video with hello world, then select the background music as favorite song, comments it and @TikTok friend.\\
\textbf{Difficulty label:} normal.\\
\textbf{The minimal step count:} 14.\\
\textbf{The app(s) involved:} TikTok.\\
\textbf{The completed form of instruction:} Open the TikTok to publish a text video with hello world, then select the background music as \{song name\}, comments it and @\{name\}.\\
\textbf{Personalized element:} favorite song, TikTok friend.\\
\textbf{Personalized info:} song name, name.
\label{box67}

\vspace{4pt}
\hrulefill
\vspace{4pt}

\textbf{Instruction ID:} 68.\\
\textbf{Natural-language instruction:} Open the bilibili to play the collected animation video and forward it to the dynamic and then @ sister.\\
\textbf{Difficulty label:} normal.\\
\textbf{The minimal step count:} 10.\\
\textbf{The app(s) involved:} Bilibili.\\
\textbf{The completed form of instruction:} Open the bilibili to play \{video name\} and forward it to the dynamic and then @\{name\}.\\
\textbf{Personalized element:} collected animation video, sister.\\
\textbf{Personalized info:} video name, name.
\label{box68}

\vspace{4pt}
\hrulefill
\vspace{4pt}

\textbf{Instruction ID:} 69.\\
\textbf{Natural-language instruction:} Open QQ and publish a text saying that the content is hello world plus a often used expression and then @ friend.\\
\textbf{Difficulty label:} normal.\\
\textbf{The minimal step count:} 12.\\
\textbf{The app(s) involved:} QQ.\\
\textbf{The completed form of instruction:} Open QQ and publish a text saying that the content is hello world plus a often \{emoji\} and then @\{name\}.\\
\textbf{Personalized element:} often used expression, friend.\\
\textbf{Personalized info:} emoji, name.
\label{box69}

\vspace{4pt}
\hrulefill
\vspace{4pt}

\textbf{Instruction ID:} 70.\\
\textbf{Natural-language instruction:} Open QQ to check the friend's QQ space, then praise the first content and comment on a often used expression.\\
\textbf{Difficulty label:} normal.\\
\textbf{The minimal step count:} 11.\\
\textbf{The app(s) involved:} QQ.\\
\textbf{The completed form of instruction:} Open QQ to check \{name\}'s QQ space, then praise the first content and comment on a \{emoji\}.\\
\textbf{Personalized element:} friend, often used expression.\\
\textbf{Personalized info:} name, emoji.
\label{box70}

\vspace{4pt}
\hrulefill
\vspace{4pt}

\textbf{Instruction ID:} 71.\\
\textbf{Natural-language instruction:} Open rednote, search for articles in the research direction, and click on the first one and click on the heart And comments are useful, plus a often used expression.\\
\textbf{Difficulty label:} normal.\\
\textbf{The minimal step count:} 11.\\
\textbf{The app(s) involved:} rednote.\\
\textbf{The completed form of instruction:} Open rednote, search for articles in the \{research direction\}, and click on the first one and click on the heart And comments are useful, plus a \{emoji\}.\\
\textbf{Personalized element:} research direction, often used expression.\\
\textbf{Personalized info:} research direction, emoji.
\label{box71}

\vspace{4pt}
\hrulefill
\vspace{4pt}

\textbf{Instruction ID:} 72.\\
\textbf{Natural-language instruction:} Open WPS Create a blank document, enter a friend's phone number, save it to the cloud disk, file name as friend, and convert to pdf format.\\
\textbf{Difficulty label:} normal.\\
\textbf{The minimal step count:} 10.\\
\textbf{The app(s) involved:} WPS.\\
\textbf{The completed form of instruction:} Open WPS Create a blank document, enter a \{phone number\}, save it to the cloud disk, file name as \{name\}, and convert to pdf format.\\
\textbf{Personalized element:} friend's phone number, friend.\\
\textbf{Personalized info:} phone number, name.
\label{box72}

\vspace{4pt}
\hrulefill
\vspace{4pt}

\textbf{Instruction ID:} 73.\\
\textbf{Natural-language instruction:} Open DianPing to search for notes on travel routes near the school and click on the first comment to publish a good comment and @ friend.\\
\textbf{Difficulty label:} normal.\\
\textbf{The minimal step count:} 11.\\
\textbf{The app(s) involved:} Dianping.\\
\textbf{The completed form of instruction:} Open DianPing to search for notes on travel routes near \{school name\} and click on the first comment to publish a good comment and @ \{name\}.\\
\textbf{Personalized element:} school, friend.\\
\textbf{Personalized info:} school name, name.
\label{box73}

\vspace{4pt}
\hrulefill
\vspace{4pt}

\textbf{Instruction ID:} 74.\\
\textbf{Natural-language instruction:} Open QQ to download files sent by friend and open and share the documents to the professional group Leave a message with a document that needs to be filled in.\\
\textbf{Difficulty label:} normal.\\
\textbf{The minimal step count:} 11.\\
\textbf{The app(s) involved:} QQ.\\
\textbf{The completed form of instruction:} Open QQ to download files sent by \{name\} and open and share the documents to \{group name\} Leave a message with a document that needs to be filled in.\\
\textbf{Personalized element:} friend, professional group.\\
\textbf{Personalized info:} name, group name.
\label{box74}

\vspace{4pt}
\hrulefill
\vspace{4pt}

\textbf{Instruction ID:} 75.\\
\textbf{Natural-language instruction:} Open DianPing to search for notes on travel routes near the school and click on the first comment to publish a good comment and @ friend and then share it with mom via text message.\\
\textbf{Difficulty label:} difficult.\\
\textbf{The minimal step count:} 18.\\
\textbf{The app(s) involved:} Dianping, Messages.\\
\textbf{The completed form of instruction:} Open DianPing to search for notes on travel routes near \{school name\} and click on the first comment to publish a good comment and @ \{name\} and then share it with \{name\} via text message.\\
\textbf{Personalized element:} school, friend, mom.\\
\textbf{Personalized info:} school name, name, name.
\label{box75}
\end{tcolorbox}

\newpage
\subsection{C.2 Human Evaluation Information}
To evaluate the effectiveness of our dataset, we conducted a human evaluation using a questionnaire. The questionnaire comprises five tasks, each assessing a distinct dimension of the PerInstruct benchmark: (i) Personalization: whether each instruction contains personalized elements; (ii) Difficulty: whether the assigned difficulty level is appropriate; (iii) Accuracy: whether the extracted personalized elements are correct and complete; (iv) Coverage: whether the instructions collectively cover typical daily mobile-usage scenarios; and (v) Diversity: whether the dataset encompasses a wide variety of applications. Detailed specifications of the human evaluation questionnaire pertaining to a single instruction are provided in Figure \ref{instruction}.
\section*{\textbf{Questionnaire}}

\renewcommand{\labelenumi}{\arabic{enumi}.}
\renewcommand{\theenumi}{\arabic{enumi}}

\tcbset{
    survey-figure/.style={
        colback=white,
        colframe=black!80,
        arc=5pt,
        boxrule=1pt,
        left=15pt, right=15pt, 
        top=10pt, bottom=10pt,
        breakable,
        enhanced,
        pad at break=5pt,
    }
}

\begin{figure}[ht] 
    \centering
    \small 
    \begin{tcolorbox}[survey-figure]
        
        \begin{center}
            \bfseries\large Questionnaire for Mobile Agent Instruction Evaluation
        \end{center}
        \vspace{6pt}
        \hrule height 0.4pt \vspace{6pt}

        \textbf{Instruction:} 
        \begin{quote}\vspace{-2pt}
            Open Baidu map, navigate to my home and then open QQ to send a message to friend that I am Almost Home.
        \end{quote}
        \vspace{2pt}
        \hrule height 0.3pt \vspace{5pt}

        \textbf{Task 1:} Do you agree that this instruction is a personalized instruction or contains personalized elements according to the following definition?
        \vspace{2pt}
        \begin{quote}\vspace{-2pt}
            \textit{Definition:} If the instruction contains words that need to be clarified by asking the user, or if certain words have different meanings for different people or devices, it can be determined that the instruction is a personalized instruction or contains personalized elements.
        \end{quote}
        \begin{enumerate}
            \item[] \makebox[1.2em][l]{1.} Strongly Disagree \hspace{5pt} 
                    \makebox[1.2em][l]{2.} Disagree \hspace{5pt} 
                    \makebox[1.2em][l]{3.} Neutral \hspace{5pt} 
                    \makebox[1.2em][l]{4.} Agree \hspace{5pt} 
                    \makebox[1.2em][l]{5.} Strongly Agree
        \end{enumerate}

        \vspace{4pt} \hrule height 0.3pt \vspace{4pt}

        \textbf{Task 2:} Do you agree with the difficulty classification of this instruction, considering the number of personalized elements, task complexity, and cross-app scope?
        \vspace{6pt}
        
        \begin{tabular}{ll}
            \textit{Difficulty:} & Difficult \\
            \textit{Minimal steps:} & 12 \\
            \textit{Personalized elements:} & 2 \\
            \textit{Cross-app:} & Yes
        \end{tabular}
        \vspace{6pt}
        
        \begin{enumerate}
            \item[] \makebox[1.2em][l]{1.} Strongly Disagree \hspace{5pt} 
                    \makebox[1.2em][l]{2.} Disagree \hspace{5pt} 
                    \makebox[1.2em][l]{3.} Neutral \hspace{5pt} 
                    \makebox[1.2em][l]{4.} Agree \hspace{5pt} 
                    \makebox[1.2em][l]{5.} Strongly Agree
        \end{enumerate}

        \vspace{4pt} \hrule height 0.3pt \vspace{4pt}

        \textbf{Task 3:} Do you think the personalized information extracted from this instruction is accurate (correct and without omissions)?
        \vspace{6pt}
        
        \textit{Personalized elements:} my home, friend
        \vspace{6pt}
        
        \begin{enumerate}
            \item[] \makebox[1.2em][l]{1.} Very Inaccurate \hspace{5pt} 
                    \makebox[1.2em][l]{2.} Inaccurate \hspace{5pt} 
                    \makebox[1.2em][l]{3.} Neutral \hspace{5pt} 
                    \makebox[1.2em][l]{4.} Accurate \hspace{5pt} 
                    \makebox[1.2em][l]{5.} Very Accurate
        \end{enumerate}

        \vspace{4pt} \hrule height 0.3pt \vspace{4pt}

        \textbf{Task 4:} After reviewing all instruction data, do you agree that this dataset can cover most scenarios of using mobile phones in real life?
        \vspace{6pt}
        
        \begin{enumerate}
            \item[] \makebox[1.2em][l]{1.} Strongly Disagree \hspace{5pt} 
                    \makebox[1.2em][l]{2.} Disagree \hspace{5pt} 
                    \makebox[1.2em][l]{3.} Neutral \hspace{5pt} 
                    \makebox[1.2em][l]{4.} Agree \hspace{5pt} 
                    \makebox[1.2em][l]{5.} Strongly Agree
        \end{enumerate}

        \vspace{4pt} \hrule height 0.3pt \vspace{4pt}

        \textbf{Task 5:} After reviewing all instruction data, do you think the variety of apps covered for a certain type of action is comprehensive enough?
        \vspace{6pt}
        
        \begin{enumerate}
            \item[] \makebox[1.2em][l]{1.} Strongly Disagree \hspace{5pt} 
                    \makebox[1.2em][l]{2.} Disagree \hspace{5pt} 
                    \makebox[1.2em][l]{3.} Neutral \hspace{5pt} 
                    \makebox[1.2em][l]{4.} Agree \hspace{5pt} 
                    \makebox[1.2em][l]{5.} Strongly Agree
        \end{enumerate}

    \end{tcolorbox}
    \caption{Detailed information on the human evaluation questionnaire.} 
    \label{instruction}
\end{figure}

\newpage
\section*{\textbf{Appendix D}\hspace{1em}\textbf{ Experimental Details}} 

\subsection{D.1 Training Details}

\textbf{Base model and Data Details.} We fine-tuned the Qwen3-8B model, specifically adapted for integration into the PerPilot framework. We used GPT-4o to generate 7,000 training samples for personalized instruction perception and personalized element extraction tasks, and 3,000 training samples for personalized information retrieval through reasoning-based exploration.

\noindent\textbf{Implementation Details.} We employ LoRA fine-tuning with a learning rate of 1e-4, cosine learning rate scheduling, an effective batch size of 16, a LoRA rank of 8, a LoRA alpha of 32, and a dropout rate of 0.1 applied specifically to the projection layers. The model is trained for 5 or 8 epochs depending on the task with a warm-up ratio of 0.05. 

\subsection{D.2 Experimental Setup}
\textbf{Test Mobile Device.} The test device employed in this experiment is a Huawei nova13 smartphone operating on Huawei HarmonyOS, and the connection between the smartphone and the computer is established through Android Debug Bridge (ADB).

\noindent\textbf{Applications Selected for Experiment.} In this experiment, the applications employed are broadly classified into five categories according to their functional purposes: Social Communication, System Tools, Life Services, Entertainment, and E-commerce, comprising a total of 27 applications. The specific list of these applications is presented in Figure~\ref{fig:D1}.

\begin{figure}[htbp]
    \centering
    \begin{tabular*}{\textwidth}{@{\extracolsep{\fill}} lllll @{}}
        \toprule
        \textbf{Social Communication} & \textbf{System Tools} & \textbf{Life Services} & \textbf{Entertainment} & \textbf{E-commerce} \\
        \midrule
        WeChat        & Phone                & Ele.me               & Bilibili               & JD.com       \\
        QQ            & Messages             & Meituan              & NetEase Cloud Music    & Taobao       \\
        Weibo         & System Calendar      & Cainiao              & TikTok                 & Pinduoduo       \\
        Rednote       & System Settings      & Didi Chuxing         &                        &              \\
                      & System Clock         & Railway 12306        &                        &              \\
                      & Browser              & Baidu Maps           &                        &              \\
                      & Weather              & Deepseek             &                        &              \\
                      & App Market           & WPS                  &                        &              \\
                      &                      & Dianping             &                        &              \\
        \bottomrule
    \end{tabular*}
    \caption{App Categories and Selected Apps}
    \label{fig:D1}
\end{figure}


\noindent\textbf{LLM Parameter Settings.}
In this experiment, the parameters configured for invoking the large language model (LLM) are specified as follows: the temperature parameter was set to 0.0 to mitigate the risk of reduced accuracy in handling personalized instructions caused by excessive model creativity; the maximum token length (max\_token) was set to 4096; and the random seed was fixed at 1234 to ensure reproducibility of experimental results.


\end{document}